\documentclass[10pt,journal]{IEEEtran}
\usepackage{amsmath,amssymb,amsfonts,mathtools}

\usepackage{algorithmic}
\usepackage{graphicx}
\usepackage{textcomp}
\usepackage{caption}
\usepackage{multirow}
\usepackage{tabularx}
\usepackage{adjustbox}
\usepackage{array}
\usepackage{epsfig} 
\usepackage{mathptmx} 
\usepackage{times} 
\usepackage[T1]{fontenc}
\usepackage{siunitx}
\usepackage{gensymb}
\usepackage{hhline}
\usepackage{accents}
\usepackage{relsize}
\usepackage{ulem}
\usepackage{balance}
\usepackage{comment}

\usepackage{hyperref}
\usepackage{tocbibind}

\newcommand{\ubar}[1]{\underaccent{\bar}{#1}}

\def\BibTeX{{\rm B\kern-.05em{\sc i\kern-.025em b}\kern-.08em
    T\kern-.1667em\lower.7ex\hbox{E}\kern-.125emX}}
\begin{document}
\title{\textcolor{black}{Efficient} Control Allocation \textcolor{black}{and 3D Trajectory Tracking} of a Highly Manoeuvrable Under-actuated Bio-inspired AUV}
\author{Walid Remmas$^*$, Christian Meurer$^*$, Yuya Hamamatsu, Ahmed Chemori, Maarja Kruusmaa 
\thanks{*These authors contributed equally to this work. \\
This work was supported in part by by the Estonian Centre of Excellence in ICT Research project EXCITE (TAR16013).}
\thanks{W. Remmas (e-mail: walid.remmas@taltech.ee) is affiliated both with the Department of Computer Systems, Tallinn University of Technology, Tallinn, Estonia, and with LIRMM, University of Montpellier, CNRS, Montpellier, France.}
\thanks{C. Meurer (e-mail: cmeurer@marum.de) is with Center for Marine Environmental Sciences (MARUM), University of Bremen, Bremen, Germany}
\thanks{A. Chemori (e-mail: ahmed.chemori@lirmm.fr) is with LIRMM, University of Montpellier, CNRS, Montpellier, France.}
\thanks{M. Kruusmaa (e-mail: maarja.kruusmaa@taltech.ee) and Y. Hamamatsu (e-mail: yuya.hamamatsu@taltech.ee) are with the Department of Computer Systems, Tallinn University of Technology, Tallinn, Estonia}
}
\maketitle

\begin{abstract}
Fin actuators can be used for both thrust generation and vectoring. Therefore, fin-driven autonomous underwater vehicles (AUVs) can achieve high maneuverability with a smaller number of actuators, but their control is challenging. This study proposes an analytic control allocation method for underactuated Autonomous Underwater Vehicles (AUVs). By integrating an adaptive hybrid feedback controller, we enable an AUV with 4 actuators \textcolor{black}{to move in 6 degrees of freedom (DOF) in simulation and up to 5-DOF in real-world experiments}. The proposed method outperformed state-of-the-art control allocation techniques in 6-DOF trajectory tracking simulations, exhibiting centimeter-scale accuracy and higher energy and computational efficiency. Real-world pool experiments confirmed the method's robustness and efficacy in tracking complex 3D trajectories, with significant computational efficiency gains (\SI{0.007}{ms} vs. \SI{22.28}{ms}). Our method offers a balance between performance, energy efficiency, and computational efficiency, showcasing a potential avenue for more effective tracking \textcolor{black}{of a large number of DOF} for under-actuated underwater robots.

\end{abstract}
\vspace{-0.5cm}
\section{Introduction}
\label{sec:introduction}

{\color{black}
Autonomous underwater vehicles (AUVs) are increasingly used in complex, dynamic environments where precise motion control is essential for task completion. Crucial to this is generating an acceleration vector in the vehicle's body frame, which governs the AUV's 3D movement and orientation along desired trajectories. This acceleration vector results from the combined forces and torques produced by the vehicle's actuators, and controlling it effectively is crucial for agile and efficient underwater motion.

Traditional AUVs typically employ fixed propellers and lifting surfaces as actuators. They manipulate the acceleration vector by adjusting the rotational velocities of the propellers, generating the necessary forces and torques along the different axes. Each actuator contributes to a specific degree of freedom (DOF), and achieving full control over all six degrees of freedom (6-DOF) usually requires at least six actuators. While this arrangement allows for precise control, it increases system complexity. 
In contrast, fin-actuated AUVs can independently change both the direction and magnitude of the generated acceleration vector through thrust vectoring. By adjusting the orientation and motion of each fin, these vehicles can produce complex force and torque patterns, controlling multiple DOF with fewer actuators, reducing system complexity and energy demands.
However, unlike fixed propellers, fins affect multiple degrees of freedom simultaneously and introduce disturbances during reorientation, complicating precise control. Additionally, fin rotation causes response delays and additional drag, further increasing the control complexity compared to propeller-driven AUVs. \\
Existing research has not fully addressed these significant challenges. Thus, the potential of fin-actuated vehicles has not been fully utilized yet. 
Specifically, state-of-the-art control for multi-DOF trajectory tracking needs to be combined with efficient and reliable control allocation, which takes the specific actuator dynamics into account. 
This paper explores the control of a fin-actuated, "turtle-like" robot with 4 fins attached to a rigid main hull capable of carrying a substantial payload \cite{dudek2007aqua, siegenthaler2013system, Salumae14}. We provide a comprehensive model of the fin actuation and derive different solutions to the control allocation problem. We further adapt and modify a state-of-the-art multi-DOF trajectory tracking controller from \cite{basso2022global}. 
Specifically, we implement the control allocation and multi-DOF controller for the fin-actuated robot U-CAT \cite{Salumae14} and enable trajectory tracking capabilities in 6-DOF in simulation and up to 5-DOF in experimental evaluations using only 4 fins.}


\color{black}

\section{Problem formulation and Related Work}
\label{sec:relatedWork}

Turtle-like fin driven vehicles have mostly been controlled manually or without feedback. A special focus was aimed at open loop gait generation \cite{konno2005development, low2007modular, zhao2008development, yao2013development, wang2012cpg} specifically focusing on gait generation using central pattern generators (CPGs) \cite{low2007modular, zhao2008development, yao2013development}. In terms of feedback based control, Geder et al. present a model-free control framework for either heading or depth control \cite{geder2013maneuvering}, and in \cite{licht2008biomimetic} attitude control for different turning maneuvers is developed. Another model-free controller for the angular rate was used by Siegenthaler et al. to stabilize forward swimming \cite{siegenthaler2013system}. All those solutions concentrated on one single DOF at a time, using simple model free control frameworks which are not robust. Chemori et al. \cite{chemori2016depth} investigated depth control for the U-CAT AUV, comparing a model-free RISE controller \cite{fischer2014nonlinear} to a standard PID control. Still concentrating on a single DOF, a robust controller shows to be effective under external disturbances.

The control frameworks described so far have been employed for set point regulation, which indicates that path following and trajectory tracking are still understudied. In \cite{plamondon2009trajectory} and \cite{plamondon2010modeling} modeling and model based control of the Aqua AUV are presented for trajectory tracking control, but again only in single DOF at a time. Multi DOF control for attitude and heave of the Auqa AUV is presented in \cite{giguere2013wide}, employing PID and PI controllers, where problems with control range and DOF coupling are avoided using gain scheduling. The approach resulted in adequate trajectory tracking, but required 45 control parameters to be tuned. Salumäe et el. \cite{salumae2017motion} proposed a framework for the U-CAT AUV, which enabled motion in several DOF (surge, yaw, heave) simultaneously. However, only heave and yaw were controlled via feedback, while surge remained an open loop control. The authors used a model based approach, termed inverse dynamics (ID), which utilized feedback linearization with acceleration feedforward \cite{fossen2011handbook}. Complexities and internal disturbances due to motion coupling were resolved by a DOF prioritization, which effectively decoupled surge from heave motions. However, this meant that heave and surge trajectories were not followed simultaneously. Meger et al. \cite{meger20143d} present a PD controller based trajectory tracking framework that extends to feedback based depth and orientation control adding roll and pitch tracking compared to \cite{salumae2017motion}, while surge tracking is still open loop and sway is not tracked at all.

While the approaches presented so far achieved satisfying results in the tested scenarios, they do not fully exploit the agility and versatility inherent in the fin-based actuation. Additionally, experimental work by Smallwood and Whitcomb \cite{smallwood2002effect} convincingly concludes that adaptive model based control approaches should be favored over fixed model based controllers.

Adaptive model based control for simultaneous tracking of 6-DOF has been thoroughly investigated and successfully implemented for standard AUVs using thruster based actuation \cite{fossen1991adaptive, antonelli2001novel, von2018stable}. However, the most common orientation representation for such frameworks relies on Euler angles, which contain singularities. An efficient singularity free orientation parametrization is needed, to derive a control framework which is provably globally stable . Unit quaternions can be used to represent 3D orientation in an efficient and singularity free manner. Quaternion based controllers have been shown to be effective for orientation control \cite{fresk2013full, louis2017quaternion}. Nevertheless, global asymptotic stabilization can not be achieved with classical continuous-time state feedback \cite{bhat2000topological}. Fjellstad and Fossen \cite{fjellstad1994singularity} introduced various discontinuous feedbacks to address the problem. However, as shown in \cite{mayhew2011quaternion} this can introduce instability to the control system. To overcome this, a hybrid feedback with a well defined switching logic can be used \cite{teel2007robust}. 
Basso et al. have theoretically and experimentally shown that a hybrid adaptive control approach can be effective for surface and underwater vehicle control \cite{basso2022global}. In their work a BlueROV2 (BlueRobotics) was used to test the control framework. The vehicle employed 8 thrusters to follow trajectories in surge, sway, heave and yaw, while regulating roll and pitch to zero. Based on those results we decided to implement the same approach for our high level control, simultaneously providing further testing results of the proposed hybrid control framework.


While several studies have addressed the control allocation problem for motion control in AUVs \cite{johansen2013control}, they have focused on propeller-based actuation. These studies have proposed various methodologies, such as direct control allocation \cite{durham1993constrained,oppenheimer2006control}, daisy chaining \cite{adams2012robust,buffington1996lyapunov}, and real-time optimization using constrained linear or quadratic programming \cite{bodson2002evaluation,paradiso1991adaptable,harkegard2002efficient,petersen2005constrained}. Additionally, recent advancements in AUV designs with tiltable thrusters have explored control allocation techniques to manage actuation redundancy \cite{jin2015six,bak2022hovering, jin2015six,bak2022hovering}.


However, control allocation methods developed for propeller-based actuation systems may not be directly applicable to fin-actuated vehicles. Unlike fixed thrusters, fins require rotation to change the direction of thrust, leading to delays and disturbances in control response. While tiltable thrusters need a similar amount of rotation, fins create a significant amount of drag when rotated. Additionally, tiltable thrusters can theoretically produce thrust throughout the rotation, while thrust generation is halted during rotation for fin-based actuation. These characteristics of fin actuation necessitate the development of tailored control allocation approaches.

In this paper we address this problem by adopting a hybrid adaptive control framework by Basso et al. \cite{basso2022global} but adapted to a fin actuated robot: instead of 8 propellers control for 4 oscillating fins is developed together with a suitable control allocation.  
In contrast to \cite{basso2022global} we also assume a possible scenario where no sway forces are available to the vehicle. To that end, we incorporated a line-of-sight-based computation \cite{breivik2009guidance} for the yaw DOF into the trajectory generation process. The line-of-sight based yaw trajectory was designed to enable sway trajectory tracking even if insufficient or no forces could be generated in the sway DOF, which is a common issue for under-actuated underwater vehicles or can be induced by actuator faults. This makes the control framework more robust and durable \textcolor{black}{at the expense of tightly coupling yaw and sway DOF}.


 \textcolor{black}{By coordinating fin movements effectively, we demonstrate that control over multiple DOF is achievable even with a reduced number of actuators. Implementing our control framework on the fin-actuated robot U-CAT \cite{Salumae14}, we achieve trajectory tracking capabilities in 6-DOF in simulations and motion control in up to 5-DOF in experimental evaluations.}
\color{black}

\begin{figure}
    \centering
    \includegraphics[width=90mm]{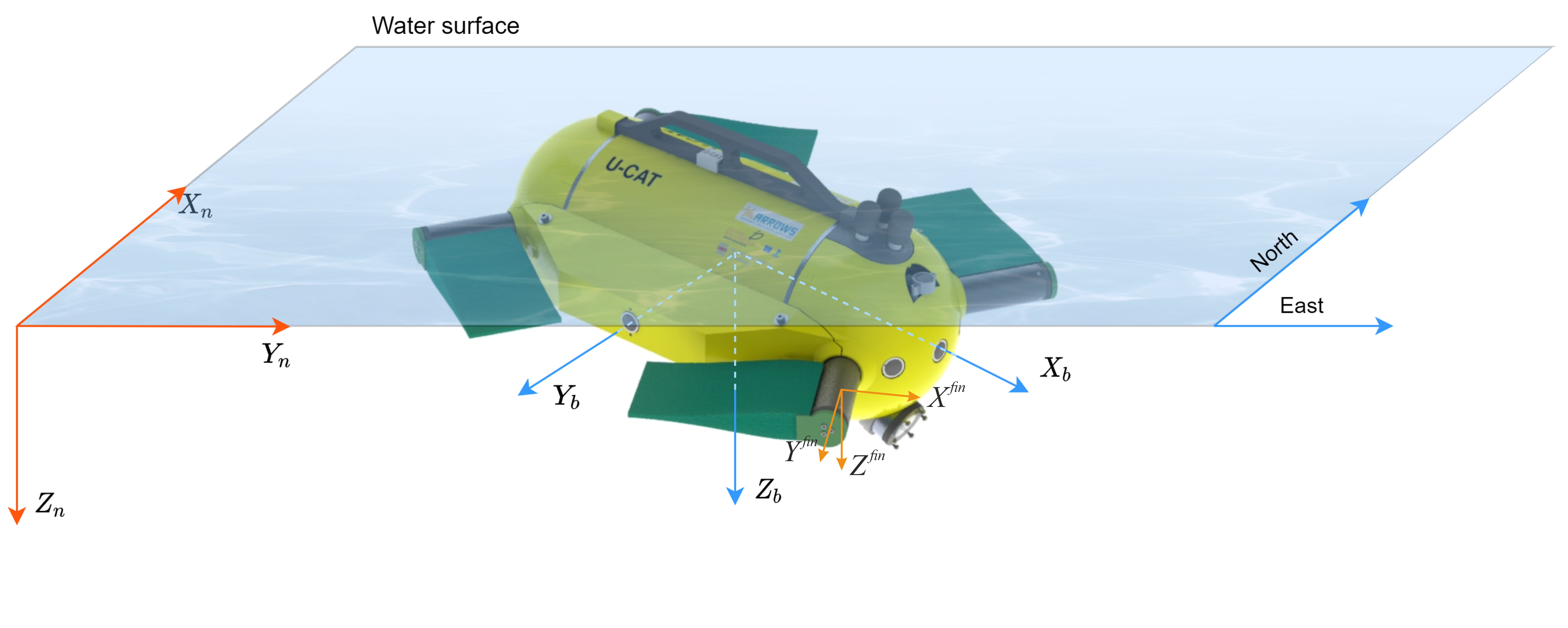}
    \captionsetup{justification=centering}
    \caption{Illustration of the earth fixed frame $R_n$ (NED convention), the body fixed frame $R_b$ and the fin rest frame $R^{fin}$ defined for the U-CAT AUV.}
    \label{fig:frames}
\end{figure}


\section{Control framework }
\label{sec:autonomy}

\begin{figure*}
    \centering
    \includegraphics[width=150mm]{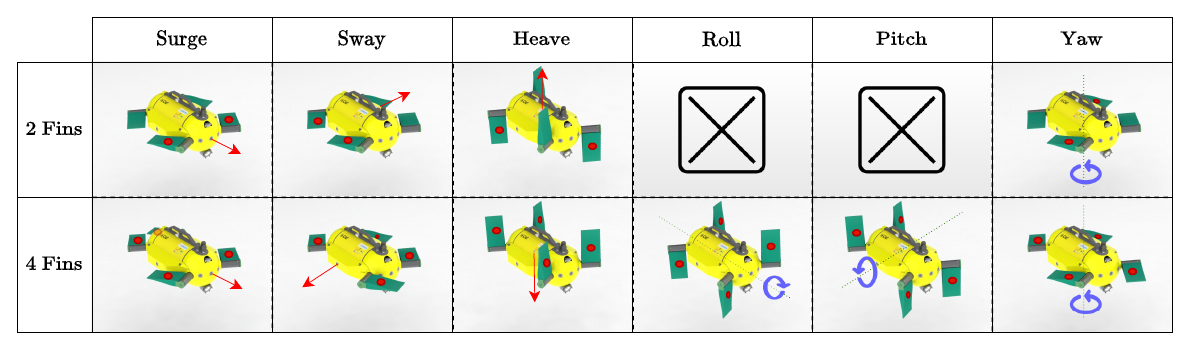}
    \caption{Illustration of the different fin configurations for controlling the U-CAT robot in each degree of freedom, using either two or four fins. The fins responsible for actuation in each configuration are marked with a red dot.}
    \captionsetup{justification=centering,margin=2cm}
    \label{fig:finConfigurations}
\end{figure*}

\subsection{Brief introduction of U-CAT}
\label{sec:UCAT}
U-CAT (shown in Fig. \ref{fig:frames}) is an autonomous four-finned underwater robot, which can be actively controlled in all 6-DOF \cite{Salumae14}.
The four motors actuating the fins are oriented as illustrated in Fig. \ref{fig:frames}. This configuration allows the robot in theory to be holonomic. \textcolor{black}{The maximally achievable thrust output is currently \SI{3.5}{N} for each fin, which will limit the achievable control authority for multi DOF control.} Further technical specifications about U-CAT are detailed in \cite{remmas2021inverse}.

\subsection{Hybrid System}

The Kinematic and Dynamic Model of the robot is described in appendix \ref{sec:appendix1}. It uses the quaternion representation to define 6-DOF by \eqref{eq:dynQuat} model where the control input is the vector of wrenches and and the output is the 6-DOF poses and velocities in the body-fixed frame. However, the map from unit quaternion to the rotation matrix describing the attitude of the vehicle defined by \eqref{eq:S3ToSO3} is non-injective and therefore creates problems in the control design, as it needs to be decided to regulate the quaternion based orientation either to $1_q$ or $-1_q$. This leads, if not considered properly, to the so called unwinding phenomenon \cite{bhat2000topological}. A commonly employed solution to this problem is the use of discontinuous feedback as shown in \cite{fjellstad1994singularity}. However, as shown in \cite{mayhew2011quaternion} this can introduce instability to the control system. Sanefelice et. al \cite{sanfelice2006robust} have theoretically and practically shown, that a discrete memory-based switching mechanism can be used to decide to which identity unit quaternion the control framework should regulate. This however, transforms the continuous time system \eqref{eq:dynQuat} into a hybrid system.

We therefore employ a hybrid system framework, where continuous and discrete evolution of the system can occur. Let $x \in \mathbb{R}^n$ be the state of a hybrid system $\mathcal{H}(f, G, C, D)$, where the continuous evolution of the states is defined by the \textit{flow map} $f: \mathbb{R}^n \to \mathbb{R}$ acting as $\dot{x} = f(x)$. In contrast, the discrete evolution of the states is defined by the \textit{jump map} $G: \mathbb{R}^n \to \mathbb{R}^n$ acting as $x^+ = G(x)$. Furthermore, the two sets $C \subset \mathbb{R}^n$ and $D \subset \mathbb{R}^n$ indicate where continuous and discrete state evolution is possible respectively. Combining the given structures provides the hybrid system: 

\begin{equation}
    \mathcal{H} = \begin{cases}
       \dot{x} = f(x) & x \in C \\
       x^+ = G(x) & x \in D
    \end{cases}
\end{equation}

Robust stability theory is available for such systems, relying on the notion of a solution to a hybrid system and requiring some regularity conditions \cite{goebel2009hybrid}. For this paper it suffices to introduce the notation given above, as we are implementing the control structure shown in \cite{basso2022global} and can thus rely on the stability proofs given therein. 


\begin{figure*}[t]
    \centering
    \includegraphics[width=150mm]{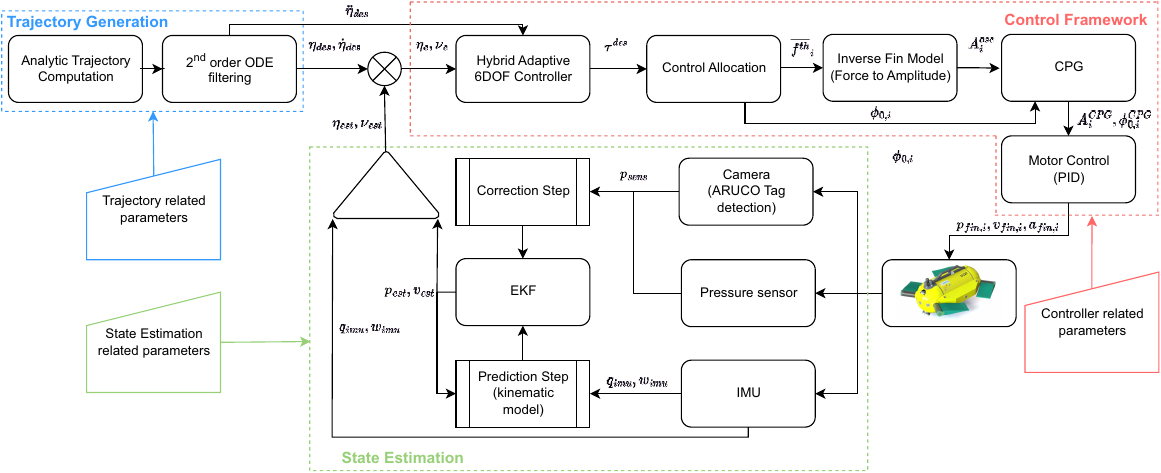}
    \captionsetup{justification=centering}
    \caption{Proposed autonomy architecture consisting of: 1) trajectory generation module in blue, 2) control module including hybrid adaptive 6-DOF controller and control allocation in red, 3) state estimation module with sensors and EKF in green.}
    \label{fig:autonomy_diagram}
\end{figure*}


\subsection{Trajectory tracking controller}
\label{subsec:control}
To show the efficacy of the proposed control allocation schemes and the autonomy framework we set up the control problem as trajectory tracking of 6-DOF. 
\subsubsection{Error System and Control Problem}

Given a desired pose $\eta_d = (p_d, q_d)$, velocity $\dot{\eta}_d$ and acceleration $\ddot{\eta}_d$, as well as the instantaneous pose $\eta = (p, q)$ and velocity $\nu$ we can then define the configuration error as:

\begin{equation}
    \eta_e = \begin{pmatrix}
        p_e \\
        q_e
        \end{pmatrix} = 
        \begin{pmatrix}
            R(q_d)^{-1} (p_d - p) \\
            q^{-1}_d \odot q
        \end{pmatrix}
\end{equation}

and the velocity error as: 

\begin{equation}
    \nu_e = \nu - \nu_r
\end{equation}

where $\nu_r = J(q_d)^\dagger \dot{\eta}_d$ can be seen as the desired velocity in the inertial frame $\dot{\eta}_d$ expressed in the vehicle frame, with $p$, $q$, $R(q)$, $J^\dagger(q)$ and the quaternion arithmetic all defined in appendix \ref{sec:appendix2}. Based on the assumption of slow rotational movement we can define the time derivative of $\nu_r$ as: 
\begin{equation}
    \dot{\nu}_r = J^\dagger(q_d) \ddot{\eta}_d
\end{equation}

The error dynamics then become:
\begin{equation}
    \mathcal{N} : \begin{cases}
        \dot{\eta}_e = \eta_e [\nu_e]_\times \\
        \dot{\nu}_e = f_e(\tau, \nu, q, q_d, \ddot{\eta}_d)
    \end{cases}
\end{equation}
where
\begin{equation}
    f_e(\tau, \nu, q, q_d, \ddot{\eta}_d) := M^{-1} (\tau - C(\nu)\nu + D(\nu)\nu + g(q)) - J^\dagger(q_d) \ddot{\eta}_d  
\end{equation}

Now the control problem can be formulated in the following way \cite{basso2022global}: For a given reference trajectory $(\eta_d, \dot{\eta}_d, \ddot{\eta}_d)$, design a hybrid feedback control law with output $\tau^{des} \in \mathbb{R}^6$ such that every solution to $\mathcal{N}$ is bounded and converges to the compact set
\begin{equation}\label{eq:controlProblem}
    \mathcal{B} = \{(\eta_e, \nu_e) ~ : ~ \eta_e \in \{p_e = 0,~ q_e = \pm 1_q\}, v_e = 0\}
\end{equation}

\subsubsection{Hybrid Adaptive Controller}
We employ an adaptive hybrid feedback control law to solve the control problem. The control law is functionally equivalent to the one presented in Basso et al. \cite{basso2022global} with slight differences in notation. Thus, we only briefly describe the used controller. For an in-depth analysis and stability proof the interested reader is referred to \cite{basso2022global} and references therein.
The control law will consist of an adaptive part based on Lyapunov design to estimate the system dynamics \cite{slotine1990hamiltonian}, of a potential function, also based on Lyapunov stability design, which includes a hysteretic switching mechanism \cite{mayhew2011quaternion}, and of a proportional feedback part acting on a reference velocity error \cite{basso2022global}. All parts are encoded by the flow and jump sets and the jump map of the resulting control system, which is essentially equivalent to a PD+ control law \cite{paden1988globally} augmented by the hysteretic switching mechanism, which leads to a hybrid control structure.  
To construct the hybrid adaptive controller we first introduce a modified reference velocity $\nu_m \in \mathbb{R}^6$ and the corresponding reference velocity error $\zeta_\nu = \nu_m - \nu_r $, to ensure convergence to the set $\mathcal{B}$. The reference velocity error is characterized by the differential equation:

\begin{equation}\label{eq:refVelDyn}
    \Lambda \dot{\zeta}_\nu = -dV_q(\eta_e) - diag(K_d) \zeta_\nu
\end{equation}

Where $K_d \in \mathbb{R}^6$ is a vector of strictly positive control gains, $\Lambda \in \mathbb{R}^{6 \times 6}$ is a diagonal matrix of strictly positive gains and $V_q(\eta_e)$ is potential function based on Lyapunov design \cite{mayhew2011quaternion}:
\begin{equation}\label{eq:potFct}
    V_q(\eta_e) = 2k (1 - h \mu_e) + \frac{1}{2} p_e^T diag(K_p) p_e
\end{equation}

with $k > 0$ and $K_p \in \mathbb{R}^3$ are strictly positive control gains. Furthermore, $\mu_e$ describes the real part of the quaternion error $q_e$ and $h$ is the hysteretic switching variable. Given the set $Q := \{-1, 1\}$ and by defining the hysteresis half-width $\varsigma \in (0,1)$ we can define the flow and jump sets of the controller by: 
\begin{align}
    C := \{(\eta_e, h) \in \mathbb{R}^3 \times \mathbb{S}^3 \times Q ~ : ~h\mu_e \geq -\varsigma\} \\
    D := \{(\eta_e, h) \in \mathbb{R}^3 \times \mathbb{S}^3 \times Q ~ : ~h\mu_e \leq -\varsigma\}
\end{align}

and the jump map is subsequently defined by: 
\begin{equation}\label{eq:jumpMap}
    G(h) = -h
\end{equation}

Using \eqref{eq:refVelDyn} the velocity error can now be redefined as: 
\begin{equation}
    \delta = \nu - \nu_m = \nu_e - \zeta_\nu
\end{equation}

which reverts to $\delta = \nu_e$ for $\zeta_\nu = 0$. Therefore, the velocity tracking objective is achieved for $(\delta, \zeta_\nu) = 0$. This error definition is suggested to be advantageous in cases where the configuration error encoded by $dV_q$ is large while the velocity error $\nu_e$ is zero \cite{basso2022global}. 

Due to uncertainties and limitations in the identified dynamics model for U-CAT and in order to create a more robust control framework, we opted to follow the approach in \cite{basso2022global} to replace the dynamics model with an adaptive on-line model identification. We assume that there exists a vector of $l$ unknown model parameters $\theta \in \mathbb{R}^l$ and a known multidimensional function of data $\Phi: \mathbb{R}^3 \times \mathbb{S}^3 \times \mathbb{R}^6 \times \mathbb{R}^6 \to \mathbb{R}^{6 \times l}$ so that \cite{fossen1991adaptive}:
\begin{equation}\label{eq:adaptiveDyn}
    M \dot{\nu}_m + C(\nu) \nu_m + D(\nu) \nu_m + g(q) = \Phi(\eta_e, \zeta, \dot{\zeta}, \delta, \nu_r, \dot{\nu}_r) \theta 
\end{equation}

The dynamics model \eqref{eq:dynQuat} can be defined with $l = 23$ parameters to be estimated, so that the restoring forces can be described by: 
\begin{equation}
   g(q) = \begin{bmatrix}
       -R(q)^T e_3 \theta_1 \\
       -[e_3]_\times R(q)^T \theta_{2:4}
   \end{bmatrix} 
\end{equation}
with $e_3 = [0, 0, 1]^T$.
Assuming symmetries in port/starboard and fore/aft directions and assuming the centre of gravity coincides with the origin of the body frame except in heave direction, the inertia matrix can be described by: 
\begin{equation}
    M = 
    \begin{bmatrix}
        \theta_5 & 0 & 0 & 0 & \theta_{11} & 0 \\
        0 & \theta_6 & 0 & -\theta_{11} & 0 & 0 \\
        0 & 0 & \theta_7 & 0 & 0 & 0 \\
        0 & -\theta_{11} & 0 & \theta_8 & 0 & 0 \\
        \theta_{11} & 0 & 0 & 0 & \theta_9 & 0 \\
        0 & 0 & 0 & 0 & 0 & \theta_{10}
    \end{bmatrix}
\end{equation}
leading to the following formulation of the matrix for Coriolis and centripetal forces:
\begin{equation}
    C(\nu) = 
    \begin{bmatrix}
        0_{3 \times 3} & [\theta_{5:7} \circ v]_{\times} - [w]_{\times} [\theta_{11} e_3]_{\times} \\
        [\theta_{5:7} \circ v]_{\times} - [\theta_{11} e_3]_{\times} [w]_{\times} & [\theta_{8:10} \circ w]
    \end{bmatrix}
\end{equation}
where $\circ$ denotes the element wise product. The hydrodynamic damping matrix can be defined by: 
\begin{equation}
    D(\nu) = -diag(\theta_{11+j}) - diag(\theta_{17+j}) \lvert \nu \rvert 
\end{equation}
with $j \in \{1, \hdots, 6\}$. The adaptive scheme is easily extendable to identify external additive disturbances as shown in \cite{basso2022global}. We chose to omit the estimation of such disturbances because we do not explicitly model the internal disturbances created by the movement of the fins, which would create instability in the parameter estimation if treated as external disturbances.

Now we define $\hat{\theta} \in \mathbb{R}^{23}$ to be the estimate of $\theta$ and $\Tilde{\theta} = \hat{\theta} - \theta$ to be the estimation error. By assuming that the parameter estimation vector $\theta$ is constant, i.e $\dot{\Tilde{\theta}} = \dot{\hat{\theta}}$ we can define a parameter update law based on Lyapunov design \cite{slotine1990hamiltonian}:
\begin{equation}\label{eq:parameterUpdate}
    \dot{\hat{\theta}} = -\Gamma^{-1} \Phi(\eta_e, \zeta_\nu, \dot{\zeta}_\nu, \delta, \nu_r, \dot{\nu_r})^T \delta
\end{equation}
where $\Gamma = diag(\gamma) \in \mathbb{R}^{23 \times 23}, \gamma >0$ is the adaptation gain matrix.

Finally, the adaptive hybrid control law can be defined by combining \eqref{eq:refVelDyn}, \eqref{eq:potFct}, \eqref{eq:jumpMap} and \eqref{eq:parameterUpdate} 
\begin{equation}\label{eq:hybridAdaptiveController}
    \begin{cases}
        \left.
        \begin{aligned}
            \dot{\zeta}_\nu &= \Lambda^{-1} (dV_q(\eta_e) + diag(K_d) \zeta_\nu) \\
            \dot{\hat{\theta}} &= Proj(-\Gamma^{-1} \Phi(\eta_e, \zeta_\nu, \dot{\zeta}, \delta, \nu_r, \dot{\nu}_r)^T\delta, \hat{\theta})
        \end{aligned}
        \right \}
        & (\eta_e, h) \in C \\
        h^+ \in G(h) & (\eta_e, h) \in D \\
        \tau^{des} = \Phi(\eta_e, \zeta_\nu, \dot{\zeta}_\nu, \delta, \nu_r, \dot{\nu}_r) \theta_a - dV_q(\eta_e) - K_d \delta 
    \end{cases}
\end{equation}

With $Proj: \mathbb{R}^{23} \times P_\varepsilon \to \mathbb{R}^{23}$ being the projection operator defined as in \cite{krstic1995nonlinear} and \cite{basso2022global}. 

\subsection{Control allocation}
\label{subsec:control allocation}

In this section, we present the control allocation process for our framework, which involves converting the desired control inputs into appropriate commands for the fin actuators. We begin by introducing the forward model of control allocation, which serves as the basis for our analysis.

Next, we discuss three different solutions to the control allocation problem (inverse models). The first solution is a naive pseudo-inverse approach, which provides a straightforward but less optimal allocation strategy. The second solution is an optimization-based approach, which aims to find an optimal distribution of control efforts using quadratic programming \cite{bodson2002evaluation}. Finally, we introduce our proposed novel control allocation which minimizes the change of fin rotations. We hypothesize that this minimization should lead to significantly less delays and minimal disturbances in control response.

We then present a model that enables the conversion of desired fin forces into oscillation amplitudes. This model takes into account the specific parameters of the fin actuators, allowing for precise control allocation.

Additionally, we incorporate a CPG algorithm, which plays a crucial role in smoothly driving the fin actuators. The CPG algorithm generates rhythmic patterns of oscillations, ensuring coordinated and synchronized movements of the fins. \\
\textbf{Forward model:}
\label{subsubsec:forwardModel}

To produce the wrenches required to control the 6-DOF body motions, U-CAT's actuation follows an oscillatory movement described by: 
\begin{equation}\label{eq:finKinematics}
    \varphi^{osc}(t) = A^{osc} ~ sin(\omega^{osc} t + \varphi^{osc}_{off}) + \phi_0
\end{equation}
with the oscillation amplitude $A^{osc}$, the oscillation rate $\omega^{osc}$, the phase offset $\varphi^{osc}_{off}$ and the zero direction of the oscillation $\phi_0$. 

To simplify the modeling, $\varphi^{osc}_{off}$ is set to zero and the instantaneous thrust $f^{th}(\varphi^{osc})$ produced by each fin is averaged over one oscillation period $T_{osc}$ \cite{ren2015hydrodynamic}: 

\begin{equation}\label{eq:thrustAveraging}
    \overline{f^{th}}(\varphi^{osc}) = \frac{1}{T_{osc}} \int^{T_{osc}}_0 f^{th}(\varphi^{osc}, \tau^{int}) ~ d\tau^{int} 
\end{equation}

with $\tau^{int}$ being the integration time variable. Given the averaged thrust of each fin, the control vector that describes the resulting wrenches in body frame $\tau$ can be derived by a concatenated frame transformation of the thrust described as a vector along the $i^{th}$ fin's zero direction $\mathbf{\overline{f^{th}}_{i}} = [\overline{f^{th}}_{i}, 0, 0, 0, 0, 0]^T $: 
\begin{equation}\label{eq:wrenchMap}
    \tau = \sum_{i=0}^{n} [Ad_{T_{i,b}}] ~ R(\phi_{0, i}) ~ \mathbf{\overline{f^{th}}_{i}}
\end{equation}

with $n$ being the total number of fins. $R(\phi_{0, i})$ is the two dimensional rotation matrix for fin $i$ which maps the thrust produced along $\phi_{0, i}$ to horizontal and vertical forces in the rest frame of the fin. 
\begin{equation}\label{eq:zeroDirectionMap}
    R(\phi_{0, i}) = 
    \begin{bmatrix}
        c \phi_{0, i} & 0 & s \phi_{0, i} & & & \\
        0 & 1 & 0 & \multicolumn{3}{c}{\smash{\raisebox{.01\normalbaselineskip}{$0_{3 x 3}$}}}\\
        -s \phi_{0, i} & 0 & c \phi_{0, i} & & & \\ 
        & & & & & & \\
        \multicolumn{3}{c}{\smash{\raisebox{.5\normalbaselineskip}{$0_{3 x 3}$}}} & \multicolumn{3}{c}{\smash{\raisebox{.5\normalbaselineskip}{$0_{3 x 3}$}}} \\
    \end{bmatrix}
\end{equation}

with $s \ast \doteq sin(\ast)$ and $c \ast \doteq cos(\ast)$.  Furthermore, $[Ad_{T_{i,b}}]$ is the adjoint representation of the homogeneous transformation matrix $T_{f, b}$ that is used to map the vertical and horizontal forces in the fins static frames to wrenches in the robot's body frame: 

\begin{equation}\label{eq:fin2BodyMap}
    [Ad_{T_{i,b}}] = 
    \begin{bmatrix}
        R(\Phi^{fin}_{i}) & 0_{3 x 3} \\
        [p^{fin}_{i}]_{\times}  R(\Phi^{fin}_{i}) &  R(\Phi^{fin}_{i})
    \end{bmatrix}
\end{equation}

with $p^{fin}_i = [x^{fin}_i, y^{fin}_i, 0]^T$ being the fin coordinates relative to the centre of the vehicle and with $ R(\Phi_{i})$ being the rotation matrix mapping from fin to body frame based on the orientation vector of the fin rest frame $\Phi^{fin}_{i} = [0, 0, \psi^{fin}_{i}]^T$ as shown in Fig. \ref{fig:frames}: 

\begin{equation}\label{eq:rotationFinBody}
    R(\Phi^{fin}) = 
    \begin{bmatrix}
        c\psi^{fin} & -s\psi^{fin} & 0 \\
        s\psi^{fin}  & c\psi^{fin}  & 0 \\
        0 & 0 & 1
    \end{bmatrix}
\end{equation}

The fin frames are arranged symmetrically (shown in Fig. \ref{fig:finConfigurations}) such that $\psi^{fin}_{1} = -\psi^{fin}_{2} - \pi = -\psi^{fin}_{3} =  \psi^{fin}_{4} + \pi$, $x^{fin}_{1} = -x^{fin}_{2} = -x^{fin}_{3} = x^{fin}_{4}$, and $y^{fin}_{1} = y^{fin}_{2} = -y^{fin}_{3} = -y^{fin}_{4}$. \\
Now by using \eqref{eq:wrenchMap} and defining $\psi^{fin} = |\psi^{fin}_{1}|$, $x^{fin} = x^{fin}_{1}$ and $y^{fin} = y^{fin}_{1}$ we can define the resulting wrenches as a system of six algebraic equations with the fins' zero directions $\phi_{0,1-4}$ and thrusts $\overline{f^{th}}_{1-4}$ as independent variables written in matrix form $AX = B$:


{\small
\begin{equation}
\label{eq:matricialForm}
\begin{bmatrix}
1 & -1 & -1 & 1 & 0 & 0 & 0 & 0\\
-1 & -1 & 1 & 1 & 0 & 0 & 0 & 0\\
0 & 0 & 0 & 0 & 1 & 1 & 1 & 1\\
0 & 0 & 0 & 0 & 1 & 1 & -1 & -1\\
0 & 0 & 0 & 0 & -1 & 1 & 1 & -1\\
-1 & 1 & -1 & 1 & 0 & 0 & 0 & 0\\
\end{bmatrix}\begin{bmatrix}
c\phi_{0,1} ~ \overline{f^{th}}_{1}\\
c\phi_{0,2} ~ \overline{f^{th}}_{2}\\
c\phi_{0,3} ~ \overline{f^{th}}_{3}\\
c\phi_{0,4} ~ \overline{f^{th}}_{4}\\
s\phi_{0,1} ~ \overline{f^{th}}_{1}\\
s\phi_{0,2} ~ \overline{f^{th}}_{2}\\
s\phi_{0,3} ~ \overline{f^{th}}_{3}\\
s\phi_{0,4} ~ \overline{f^{th}}_{4}\\
\end{bmatrix} = \begin{bmatrix}
\frac{\tau_x}{c\psi^{fin}} \\
\frac{\tau_y}{s\psi^{fin}} \\
\tau_z \\
\frac{\tau_{\Phi}}{y^{fin}} \\
\frac{\tau_{\Theta}}{x^{fin}} \\
\frac{\tau_{\Psi}}{M_a}
\end{bmatrix}
\end{equation}
}
with $M_a = x^{fin} ~ s\psi^{fin} - y^{fin} ~ c\psi^{fin}$. \\


\textbf{Inverse model:}
\label{subsubsec:inverseModel}

Here, we present the three studied control allocation solutions that allows to control the robot in 6-DOF simultaneously.

\subsubsection{Direct solution}
Since the matrix $A$ is of full rank, a non-unique solution for the system using the Moore-Penrose inverse exists, hence $X=A^T(AA^T)^{-1} B$. Given the aforementioned symmetric configuration of the $i = 1, ..., 4$ fins, the solution can be expressed in the following form, where fin forces and orientations are still coupled:

\begin{align}
    c\phi_{0,i} ~ \overline{f^{th}}_{i} & = \frac{1}{4}\left(\frac{\tau_x}{c\psi^{fin}_{i}} + \frac{\tau_y}{s\psi^{fin}_{i}} + \frac{sign(\psi^{fin}_{i}) ~ \tau_\Psi}{M_a}\right) \label{eq:inverse_coupled_1} \\
    s\phi_{0,i} ~ \overline{f^{th}}_{i} & = \frac{1}{4}\left(\tau_z + \frac{\tau_\Phi}{y^{fin}_{i}} - \frac{\tau_\Theta}{x^{fin}_{i}}\right) \label{eq:inverse_coupled_2}
\end{align}

By defining the sums of wrenches in \eqref{eq:inverse_coupled_1} and \eqref{eq:inverse_coupled_2} in terms of horizontal $f^{hor}_{i}$ and vertical $f^{vert}_{i}$ contributions respectively as:

\begin{align}
    f^{hor}_{i} & = \frac{\tau_x}{c\psi^{fin}_{i}} + \frac{\tau_y}{s\psi^{fin}_{i}} + \frac{sign(\psi^{fin}_{i}) ~ \tau_\Psi}{M_a} \label{eq:horizontalWrenches}\\
    f^{ver}_{i} & = \tau_z + \frac{\tau_\Phi}{y^{fin}_{i}} - \frac{\tau_\Theta}{x^{fin}_{i}} \label{eq:verticalWrenches} 
\end{align}

and then dividing equation \eqref{eq:inverse_coupled_2} by equation \eqref{eq:inverse_coupled_1}, we can deduce the zero direction $\phi_{0,i}$ for each fin as follows: 

\begin{equation}\label{eq:inversePhi}
    \phi_{0,i} = \arctan2{\left(f^{vert}_{i},f^{hor}_{i}\right)}
\end{equation}

The thrust forces required by each fin are then derived by squaring and adding equations \eqref{eq:inverse_coupled_1} and \eqref{eq:inverse_coupled_2}: 

\begin{equation}\label{eq:inverseF}
    \overline{f^{th}}_{i} = \frac{1}{4} ~ \sqrt{(f^{hor}_{i})^2 + (f^{vert}_{i})^2}
\end{equation}

\textcolor{black}{which is equivalent to \textit{extended thrust representations} in the literature \cite{fossen2006survey}}. For this control allocation method, denoted as CA$_{inv}$, it is important to note that all four fins need to be actuated regardless of the controller's output $\tau^{des}$. 
This characteristic can lead to undesired fin rotations, such as all four fins rotating $\SI{180}{\degree}$ when the surge component changes its sign. These rotations result in significant disturbances caused by the fin movements when changing the fin's orientation.
\\

\subsubsection{Optimization based solution}
An alternative approach to solving the control allocation problem involves employing near real-time optimization techniques. In this optimization-based solution, the constraints presented in equation \eqref{eq:matricialForm} are reconfigured \textcolor{black}{to avoid the explicit inclusion of trigonometric functions in the system’s equations or the constraints}:

\begin{equation}
    \min _{\overline{f^{th}}}~J =  \overline{f^{th}}^T \overline{f^{th}}
\end{equation}

Subject to: \\
{\small
\begin{align}
 \begin{split}
 \tau_x &= c\psi_f \left(\Gamma^{opt}_1 ~ \overline{f^{th}}_{1} - \Gamma^{opt}_2 ~ \overline{f^{th}}_{2} - \Gamma^{opt}_3 ~\overline{f^{th}}_{3} + \Gamma^{opt}_4 ~ \overline{f^{th}}_{4} \right) \\
  \tau_y &= s\psi_f \left(-\Gamma^{opt}_1 ~ \overline{f^{th}}_{1} - \Gamma^{opt}_2 ~ f_{th,2} + \Gamma^{opt}_3 ~ f_{th,3} + \Gamma^{opt}_4 ~ \overline{f^{th}}_{4} \right)\\
  \tau_z &= \Lambda^{opt}_1 ~ \overline{f^{th}}_{1} + \Lambda^{opt}_2 ~ \overline{f^{th}}_{2} + \Lambda^{opt}_3 ~ \overline{f^{th}}_{3} + \Lambda^{opt}_4 ~ \overline{f^{th}}_{4}\\
  \tau_{\Phi} &= y_f \left(\Lambda^{opt}_1 ~ \overline{f^{th}}_{1} + \Lambda^{opt}_2 ~\overline{f^{th}}_{2} - \Lambda^{opt}_3 ~ \overline{f^{th}}_{3} - \Lambda^{opt}_4 ~ \overline{f^{th}}_{4} \right)\\
  \tau_{\Theta} &= x_f \left(-\Lambda^{opt}_1 ~ \overline{f^{th}}_{1} + \Lambda^{opt}_2 ~ \overline{f^{th}}_{2} + \Lambda^{opt}_3 ~ \overline{f^{th}}_{3} - \Lambda^{opt}_4 ~ \overline{f^{th}}_{4} \right)\\
  \tau_{\Psi} &= M_a \left(-\Gamma^{opt}_1 ~ \overline{f^{th}}_{1} + \Gamma^{opt}_2 ~ \overline{f^{th}}_{2} - \Gamma^{opt}_3 ~ \overline{f^{th}}_{3} + \Gamma^{opt}_4 ~ \overline{f^{th}}_{4} \right)\\
  1 &= \Gamma^{opt^2}_{i} + \Lambda^{opt^2}_{i} \qquad i = 1 \hdots 4 \\
  -1 &\leq \Gamma^{opt}_i \leq 1 \qquad i = 1 \hdots4 \\
  -1 &\leq \Lambda^{opt}_i \leq 1 \qquad i = 1 \hdots 4 \\
  0 &\leq \overline{f^{th}}_{i} \leq F_{max} \qquad i = 1 \hdots 4
  \end{split}
\end{align}
}
with $\overline{f^{th}} = [\overline{f^{th}}_{1}, \overline{f^{th}}_{2}, \overline{f^{th}}_{3}, \overline{f^{th}}_{4}]^T$ aggregating the average thrusts for each individual fin

We have defined the system in this particular way to explicitly solve for the required forces and zero-directions, even though at this stage, the cost function's objective is to minimizes the thrust forces only. It is worth noting that the optimization process does not include an additional term to minimize the zero-direction change, as it was found to be counterproductive and led to non-convergence of the algorithm.

The optimization problem is solved during runtime to find the optimal parameters $[\overline{f^{th}}, \Gamma^{opt}, \Lambda^{opt}]$ using Sequential Quadratic Programming (SQP) \cite{sqp}. The initial conditions for all parameters are set to $0$. The zero-directions $\phi_0$ are then computed using the following formula:
\begin{equation}
    \phi_{0,i} = atan2(\Lambda^{opt}_i, \Gamma^{opt}_i) \qquad i = 1 \hdots 4
\end{equation}

This optimization-based solution for control allocation presents certain limitations. Firstly, it does not explicitly include the minimization of zero-direction change in its objective cost function, which will not guarantee finding the optimal solution for minimum zero-direction change. Secondly, using an iterative optimization algorithm requires more computation resources and capabilities to ensure timely execution within the hardware constraints.

Throughout the rest of the paper, the optimization-based control allocation method will be referred to as CA$_{opt}$.\\

\subsubsection{Proposed analytic solution}

To address the limitations of the two control allocation methods discussed, namely CA$_{inv}$ and CA$_{opt}$, we propose a novel approach that analytically solves the control allocation problem while considering the minimization of the zero-direction change.

Taking advantage of the symmetrical configuration of U-CAT's fins, we observe that for certain degrees of freedom such as surge, sway, heave, and yaw, it is possible to only employ two fins to provide a thrust vector to move in certain directions, as illustrated in Fig. \ref{fig:finConfigurations}. A key insight here is, that forces in opposing directions can be generated with opposing fins, which can be mathematically encoded with Heaviside unit step functions $H(\cdot)$ \cite[p.~61]{bracewellHeaviside}. The Heaviside functions are basically used to determine which set of 2 fins is needed at any point in time to generate a desired force in a specific direction.

In the following, we derive the control allocation equations \eqref{eq:inversePhi} and \eqref{eq:inverseF} as functions of the desired numbers $n_d$ of fins used for the surge, sway, heave and yaw degrees of freedom. In the current setup $n_d$ for each DOF is static and defined a priori.  We re-define the sums of wrenches from equations \eqref{eq:horizontalWrenches} and \eqref{eq:verticalWrenches} as:

\begin{align}
    f^{hor}_{i}(\tau, n_d) & = \frac{h_H(\tau_x, c\psi^{fin}_{i}, n_d)}{c\psi^{fin}_{i}} + \frac{h_H(\tau_y, s\psi^{fin}_{i}, n_d)}{s\psi^{fin}_{i}} \\
    & + \frac{h_H(\tau_\Psi, \psi^{fin}_{i}, n_d)}{M_a} \label{eq:horizontalWrenchesVariable} \\
    f^{vert}_{i}(\tau, n_d) & = h_H(\tau_z, -\psi^{fin}_{i}, n_d) + \frac{\tau_\Phi}{y^{fin}_{i}} - \frac{\tau_\Theta}{x^{fin}_{i}} \label{eq:verticallWrenchesVariable}
\end{align}

with $h_H(\cdot)$ determining the fin usage through Heaviside functions as:

\begin{equation}\label{eq:heavihelper}
    h_H(\tau, s , n_d) = \begin{cases} 
        2 ~ H\left(sign(s)\tau\right) ~ \tau & n_d = 2 \\
        \tau & n_d = 4
    \end{cases}
\end{equation}

By inserting \eqref{eq:horizontalWrenchesVariable}, \eqref{eq:verticallWrenchesVariable} in \eqref{eq:inversePhi} and \eqref{eq:inverseF}  the zero direction and thrust force for each fin can be described by:

\begin{align}
    \phi_{0,i} & = \arctan{\left(\frac{f^{ver}_{i}(\tau, n_d)}{f^{hor}_{i}(\tau, n_d)}\right)} \label{eq:inversePhiVariable} \\
    \overline{f^{th}}_{i} & = \frac{1}{4} ~ \sqrt{f^{hor}_{i}(\tau, n_d)^2 + f^{ver}_{i}(\tau, n_d)^2} \label{eq:inverseFVariable}
\end{align}

This proposed control allocation scheme described \textcolor{black}{by \eqref{eq:inverseFVariable}} has several advantages. It allows the robot to have more flexibility of movement. 
\color{black}

\begin{figure}
    \centering
    \includegraphics[width=50mm]{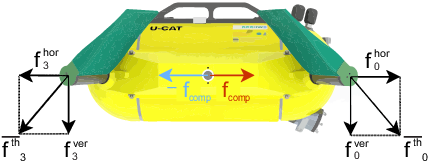}
    \caption{Illustration of the horizontal force compensation principle to minimize the fins' zero-direction change when controlling vertical forces.}
    \captionsetup{justification=centering,margin=2cm}
    \label{fig:forceCompensation}
\end{figure}

Moreover, we can again exploit the fins' symmetrical configuration to minimize the change in zero direction control. Indeed, when controlling forces and torques around the vertical plane, then the fins are oriented with an angle of $\pm \frac{\pi}{2}$ according to equation \eqref{eq:inversePhiVariable}. To avoid the non-linearity in the $\arctan$ function, we can introduce two opposing horizontal forces that are naturally compensated, as illustrated in Fig. \ref{fig:forceCompensation}. We add a term $f_{comp}$ such that:

\begin{align}\label{eq:inverseFVariableComp}
    \overline{f^{th}}_{i} & = \frac{1}{4} ~ \sqrt{\left(f_{comp} + f^{hor}_{i}(\tau, n_d)\right)^2 + f^{vert}_{i}(\tau, n_d)^2}  \\
    \phi_{0,i} & = \arctan{\left(\frac{f^{ver}_{i}(\tau, n_d)}{f_{comp} + f^{hor}_{i}(\tau, n_d)}\right)}
\end{align}

with 

\begin{align}\label{eq:fcomp}
    f_{comp} &= \alpha_{comp} \sum_j(1-f_{j}^{norm}) |f_{j}^{norm}| \\
    f_{j}^{norm} &= \frac{|\tau_j|}{\overline{f^{th}}_{max}}
\end{align}

with $j=z,\Theta,\Psi$ and where $\overline{f^{th}}_{max}$ is the maximum producible thrust or torque for each fin within specific limits for fin oscillation frequency and amplitude.
This means that the compensation is not introduced in two cases: when $f_{j}^{norm} = 0$ and when $f_{j}^{norm} = 1$. The parameter $\alpha_{comp}$ is a scalar that weighs the effect of the the introduced term $f_c$ to minimize the zero-direction change, at the expense of higher oscillation amplitudes. 


The proposed control allocation method will be referred to as CA$_{prop}$.
\\

\subsection{Force to amplitude}
Once the required thrust $\overline{f^{th}}_{i}$ for each fin is computed, it is converted to a fin-oscillating amplitude using the inverse model described in \cite{remmas2021inverse} as follows:

\begin{equation}
\label{eq:force2amplitude}
A^{osc}_{i} =\arccos \left( \frac{-{\overline{f^{th}}_{i}}}{2 C_d \rho S_f (r_c\omega^{osc})^2}+1  \right)
\end{equation}

where $\omega^{osc}$ denotes the angular rate of the fin, $\rho$ is the water density, $r_c$ is the distance between the rotation axis and the center of gravity of the fin, $C_d$ stands for the drag coefficient, and $S_f$ is the projection area of the fin. The resulting amplitudes $A^{osc}_i$ and zero-directions $\phi_{0,i}$ are then filtered using a Central Pattern Generator (CPG) algorithm. 

\subsection{CPG algorithm}
\label{subsubsec:CPG}
The CPG is used to ensure smooth and continuous transitions, which significantly reduces the effect of the non-modelled fin lateral forces. The equations of the CPG used in our study are adopted from \cite{sproewitz2008learning}:

\begin{align}
\dot{\zeta}^{CPG}_{i} & = \omega^{osc}_i \\
    \ddot{A}^{CPG}_{i} & = K_{amp} \left( \frac{K_{amp}}{4} (A^{osc}_i - A^{CPG}_{i}) - \dot{A}^{CPG}_{i} \right) \\
    \ddot{\phi}^{CPG}_{0,i} & = K_{zd} \left( \frac{K_{zd}}{4} ({\phi}_{0,i} - \phi^{CPG}_{0,i}) - \dot{\phi}^{CPG}_{0,i} \right) \\
    \phi^{CPG}_{i} & = {\phi}^{CPG}_{0,i} + {A}^{CPG}_{i} cos({\zeta}^{CPG}_{i})
\end{align}

where $\phi^{CPG}_{i}$ is the zero-direction angle (in radians) extracted from the oscillator and $\zeta^{CPG}_i$, $A^{CPG}_{i}$ and $\phi^{CPG}_{0,i}$ are state variables that encode the phase, amplitude and the zero-direction offset of the oscillations (in radians), respectively. The parameters $\omega^{osc}_i$, $A^{osc}_i$ and $\phi_{0,i}$ are control parameters for the desired angular rate, amplitude and offset of the oscillations generated by the control allocation.

\section{Numerical simulations}
\label{sec:simulations}

\begin{figure}[b!]
    \centering
    \includegraphics[width=50mm]{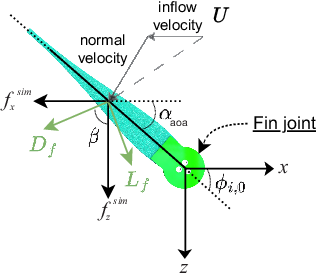}
    \caption{Visualization of forces on $i^{th}$ fin-based on a simple lift and drag model (model and figure adopted from \cite{georgiades2009simulation}).}
    \captionsetup{justification=centering,margin=2cm}
    \label{fig:fin_lift_drag}
\end{figure}
A simulation framework allows evaluating the presented autonomy framework, with a specific focus on controller and control allocation. The simulation is purely written in Python for fast deployment and is based on Fossen's vectorial dynamics model \eqref{eq:dynQuat}. The numerical values for the dynamic model parameters are denoted in the 
\href{https://drive.google.com/drive/folders/1UonxM-o0iKqdTh8wGTRgRzLZNU6QQkQP}{\textit{\underline{linked online document}}}. 
Furthermore, a lift and drag based model \cite{healey1995toward} is utilized to simulate the forces produced by the fins. Trajectory generation is described in Appendix \ref{sec:appendix2} and the state estimation is described in Appendix \ref{sec:appendix3}.   Control is simulated according to the descriptions shown in the previous sections. 
To attain robust and reliable estimates of the performance of our proposed control allocation solution in comparison to the presented state-of-the-art solutions $CA_{prop}$ and $CA_{inv}$ solutions we use Monte Carlo Simulations in combination with a set of assessment metrics. Those metrics are used to estimate tracking performance as well as, physical and computational efficiency. All simulations were run on a laptop with an 11$^{th}$ generation Intel Core i7-1165G7 processor and \SI{16}{GB} RAM, running Ubuntu 20.04.6 LTS.  

We simulated two scenarios, full 6-DOF trajectory tracking (6T) and 3 DOF trajectory tracking with roll and pitch stabilization (3T2S). In the 6T scenario we assumed that the robot is fully actuated, while in the 3T2S scenario we assumed, that actuation in the sway direction is not efficient enough. The lack of available sway force makes the system underactuated and we implement a lookahead modification of the generated yaw trajectory to ensure that the robot can still follow the looped trajectories without the need for sway forces. The full description of the generated trajectories can be found in appendix \ref{sec:appendix2}. 

The following paragraphs detail the structure of the simulation framework, including the models used to simulate the dynamics of the robot and fins, the Monte Carlo setup including modeled sensor noise, and the methodology for tuning the control system's hyper-parameters.\\

\subsection{Dynamics simulation}
\label{subsec:finModelling}
U-CAT's motion dynamics were simulated using Fossen's hydrodynamic model presented in \eqref{eq:dynQuat}. The differential equations derived from this mode were solved using the $4^{th}$ order Runge-Kutta algorithm with a step size of \SI{0.01}{s}. This approach enabled the accurate representation of U-CAT's behavior in a controlled underwater environment, taking into account the various hydrodynamic parameters and the effect of the fins' oscillations. The relevant parameters were identified with an approach described in \cite{salumae2017motion}.

The dynamic model used to simulate the fins of U-CAT was based on the rigid paddle model \cite{healey1995toward}. The efficacy of this model was demonstrated and validated in previous research \cite{georgiades2009simulation} when simulating the underwater hexapod robot AQUA \cite{georgiades2004aqua}.

The fins of U-CAT are simulated to generate horizontal $f^{sim}_x$ and vertical $f^{sim}_z$ forces relative to the fin's rest frame, as depicted in Fig.~\ref{fig:fin_lift_drag} through the following expressions:

\begin{align}
\begin{split}
f^{sim}_x &= D_f \sin(\beta) + L_f \cos(\beta) \\
f^{sim}_z &= -L_f \sin(\beta) + D_f \cos(\beta)
\end{split}
\end{align}

Where $\beta$ represents the direction of flow impinging on the fin, and $D_f$ and $L_f$ denote the lift and drag forces, respectively. These forces are defined as:

\begin{align}\label{eq:FinLiftDrag}
\begin{split}
L_f &= 0.5 \rho U_f^2 S_f C_{Lmax} \sin(2 \alpha_{aoa}) \\
D_f &= 0.5 \rho U_f^2 S_f C_{Dmax} (1 - \cos(2 \alpha_{aoa}))
\end{split}
\end{align}

where $\alpha_{aoa}$ is the angle of attack, and $U_f$ is the velocity of the flow impacting the fin. Additionally, $C_{Lmax}$ and $C_{Dmax}$ are the paddle's maximum lift and drag coefficients, respectively, over the full $360^\circ$ range of the angle of attack. The coefficients were tuned so that the resulting simulated thrust output was quantitatively similar to the output from laboratory experiments shown in \cite{salumae2017motion}. 

Finally, the resulting simulated wrenches in the body frame can be computed by:
\begin{equation}\label{eq:WrenchSim}
    \tau^{sim} = \sum_{m=0}^{n} [Ad_{T_{m,b}}] ~ \mathbf{f^{sim}_m} 
\end{equation}
with $\mathbf{f^{sim}} = [f^{sim}_x, 0, f^{sim}_z, 0, 0, 0]^T$ being the vector of fin forces in the fin's rest frame, and $[Ad_{T_{m,b}}]$ defined as in \eqref{eq:fin2BodyMap}. For a more in-depth discussion of the simulation of fin forces, the reader is encouraged to consult \cite{georgiades2009simulation}.\\

\subsection{Monte Carlo Framework}
\label{subsec:MonteCarloFramework}
\begin{figure}[b!]
    \centering
    \includegraphics[width=0.8\linewidth]{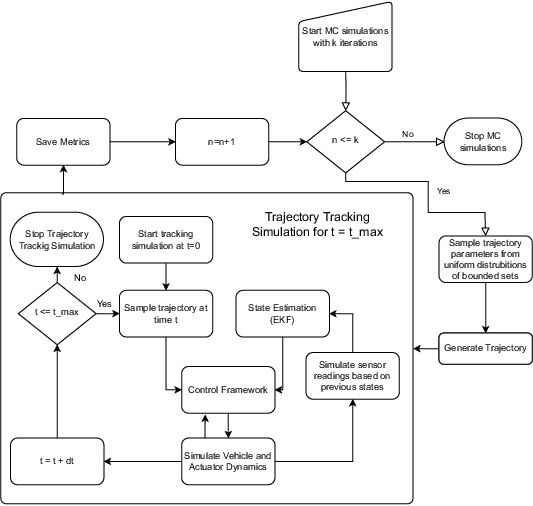}
    \caption{Flowchart for Monte Carlo Simulations.}
    \label{fig:MCSim}
\end{figure}

We ran 500 iterations in a Monte Carlo simulation framework for the 3 studied control allocation methods, namely CA$_{inv}$, CA$_{opt}$ and CA$_{prop}$, while keeping the other modules of the autonomy framework constant.
In each run of the Monte Carlo simulations, a trajectory scenario (6T or 3T2S) and type (Ellipse or Lissajous) are randomly picked. The relevant parameters for the specific trajectory are uniformly sampled from a bounded set as shown in Table \ref{tab:trajectory-parameters}. Each trajectory tracking simulation is run for \SI{400}{s}. 
Additionally, to ensure a more realistic trajectory tracking evaluation, we simulated the sensor suite shown in Fig. \ref{fig:autonomy_diagram}, including an IMU for orientation estimation, a pressure sensor for depth estimation and a camera-based position estimator. For all sensors, the true simulated state of the robot was used as a basis, and white Gaussian noise (WGN) with a zero mean was added. The variances for the WGN additions are chosen to reflect the sensor characteristics of the real vehicle used in subsequent experiments and are also shown in Table \ref{tab:trajectory-parameters}. The controller gains were kept constant for each simulation run. The flowchart in Fig. \ref{fig:MCSim} describes the full setup of the Monte Carlo simulation framework.

\begin{table}
\centering
\caption{Permissible ranges of trajectory generation and sensor simulation parameters for Monte Carlo Simulations.}
\label{tab:trajectory-parameters}
\begin{tabular}{c|cc|}
\cline{2-3}
                                 & \multicolumn{1}{c|}{\textbf{Ellipsoidal}} & \textbf{Lissajous} \\ \hline
\multicolumn{3}{|c|}{Trajectory generation} \\ \hline
\multicolumn{1}{|c|}{$[x_{0}, A_{x}, \omega_x]$}    & \multicolumn{2}{c|}{[0.3, 0.5 - 2.5, 0.01 - 0.05]}                     \\ \hline
\multicolumn{1}{|c|}{$l_x$} & \multicolumn{1}{c|}{/} & 0.5 - 2.0           \\ \hline
\multicolumn{1}{|c|}{$[A_{y}, \omega_y]$}    & \multicolumn{2}{c|}{[0.5 - 2.5, 0.01 - 0.05]}            \\ \hline
\multicolumn{1}{|c|}{$l_y$} & \multicolumn{1}{c|}{/} & 0.5 - 2.0           \\ \hline
\multicolumn{1}{|c|}{$[A_{z}, \omega_z, c_\phi]$}    & \multicolumn{2}{c|}{[0.1 - 0.5, 0.01 - 0.05, 0.05 - 0.15]}            \\ \hline
\multicolumn{1}{|c|}{$t^*$}    & \multicolumn{2}{c|}{0.1}                     \\ \hline
\multicolumn{1}{|c|}{$[\gamma_1, \gamma_2]$} & \multicolumn{2}{c|}{$[7.0_{6 \times 1}, 1.0_{6 \times 1}]$}                     \\ \hline
\multicolumn{3}{|c|}{Sensor simulation} \\ \hline
\multicolumn{1}{|c|}{$\sigma^{WGN}_{pose}$} & \multicolumn{2}{c|}{$[0.002, 0.002, 0.002, 0.0017, 0.0017, 0.0017]^T$}                     \\ \hline
\multicolumn{3}{|c|}{State estimation} \\ \hline
\multicolumn{1}{|c|}{$\sigma_Q$} & \multicolumn{2}{c|}{$[0.01, 0.01, 0.01, 0.02, 0.02, 0.01]^T$} \\ \hline
\multicolumn{1}{|c|}{$\sigma_R$} & \multicolumn{2}{c|}{$[0.002, 0.002, 0.002]^T$}  \\ \hline

\end{tabular}
\end{table}

\subsection{Data Analysis}
\label{subsec:DataAnalysisSim}
We defined several evaluation metrics, inspired by suggestions of Manhaes et al \cite{manhaes2017metrics}, to characterize the performance of our proposed control framework. Based on the tracking error with the orientation error represented by Euler angles $\prescript{\Phi}{}\eta_{e, j} = [p_{e, j}, \Phi_{e, j}]^T$ the following evaluation metrics were defined and used: 

\begin{itemize}
    \item \textit{Root Mean Squared Error} between desired and estimated robot trajectory for linear and angular DOF [$\mathbf{RMSE_{lin}}$ and $\mathbf{RMSE_{ang}}$]
    \begin{equation}
        RMSE = \sqrt{\frac{\displaystyle\mathlarger{\sum_{j=1}^N} \prescript{\Phi}{}\eta_{e, j}^T \prescript{\Phi}{}\eta_{e, j}}{N}}
    \end{equation}
    \item \textit{Maximum Error Magnitude} between desired and estimated robot trajectory for linear and angular DOF [$\mathbf{MEM_{lin}}$ and $\mathbf{MEM_{ang}}$]
    \begin{equation}
        MEM = \max\left\{\left\lVert \prescript{\Phi}{}\eta_{e, 1} \right\rVert, \left\lVert \prescript{\Phi}{}\eta_{e, 2} \right\rVert, \dots, \left\lVert \prescript{\Phi}{}\eta_{e, N} \right\rVert\right\}
    \end{equation}
    \item \textit{Mean Absolute Wrench} [$\mathbf{MAW}$]
    \begin{equation}
        MAW = \frac{1}{N} \displaystyle\mathlarger{\sum_{n=1}^N} \left\lVert \frac{\tau^{sim}_n}{corr_{Ma}}\right\rVert
    \end{equation}
    \item \textit{Maximum Wrench} [$\mathbf{MW}$]
    \begin{equation}
        MW = \max\left\{\left\lVert \frac{\tau^{sim}_1}{corr_{Ma}}\right\rVert, \left\lVert \frac{\tau^{sim}_2}{corr_{Ma}}\right\rVert, \dots, \left\lVert \frac{\tau^{sim}_n}{corr_{Ma}}\right\rVert\right\}
    \end{equation}
    \item \textit{Median Computation Time} for control allocation [$\mathbf{MCT}$]
    \begin{equation}
        MCT = med\left\{dt^{ca}_1, dt^{ca}_2, \dots, dt^{ca}_N\right\}
    \end{equation}
    \item \textit{Mean Allocation Error} for linear and angular degrees of freedom [$\mathbf{MAE_{lin}}$ and $\mathbf{MAE_{ang}}$]
    \begin{equation}
        MAE = \frac{1}{N} \displaystyle\mathlarger{\sum_{j=1}^N} \left\lvert \tau^{des}_j - \tau^{sim}_j \right\rvert
    \end{equation}
\end{itemize}

Here, $N$ is defined as the number of data points for a specific simulation or experimental run. Furthermore, $\tau^{des}_j$ describes the commanded wrench, which is produced by the hybrid adaptive controller \eqref{eq:hybridAdaptiveController}, whereas $\tau^{sim}_j$ describes the wrenches produced by the simulation of the oscillating fins \eqref{eq:WrenchSim}. To correctly represent the wrench-related metrics, the torques in the simulated wrenches $\tau^{sim}$ are converted into forces using the respective moment arms $corr_{Ma} = [0, 0, 0, x^{fin}, y^{fin}, M_a]^T$. The time it takes, given $\tau^{des}_j$, for the respective control allocation algorithm to assign the fin kinematics is described by $dt^{ca}_j = t^{ca}_{0, j} - t^{ca}_{j}$.

RMSE and MEM describe average and maximum errors and thus an estimate of tracking performance, while MEM and MW provide a measure of energy consumption in terms of physical action demanded by the tested allocation frameworks. MCT, in turn, provides a measure of computational energy demanded by the allocation algorithms. Finally, MAE can be used to evaluate the accuracy of the tested allocation algorithms. 
All evaluation metrics are computed for each single simulation run in the Monte Carlo framework, and then the median with interquartile range (IQR) is computed over all runs to provide a robust estimate of overall performance.

\subsection{Hyper-parameter tuning}
This section describes the hyper-parameter tuning process in numerical simulations for all the autonomy modules described in Section \ref{sec:autonomy}, as well as, appendices \ref{sec:appendix2} and \ref{sec:appendix3}.
The trajectory generator's coefficients $\gamma_1$ and $\gamma_2$ were manually adjusted to ensure that the generated velocities and accelerations fall within the physically feasible range for the robot. The gain matrix $\Lambda = I_6 $ was defined according to \cite{basso2022global}. The number of fins involved in force / torque generation for each DOF $n_d$ was manually chosen to minimize the required fin rotations to generate the desired forces. The boundaries and boundary layers for the adaptive model estimation were defined to keep each variable within a range that is still physically plausible. To define the starting parameters of the adaptive part of the controller, model parameters identified in previous work \cite{salumae2017motion} were taken as guiding values. However, to reflect the uncertainty in the identification process and to better estimate the quality of the adaptive part of the controller, parameters provided by model identification \cite{salumae2017motion} were rounded and decreased to base values like $10.0$, $100.0$ or $5.0$. Finally, the adaptation gain matrix $\Gamma$ was tuned manually based on simulation results. The numeric values for each parameter can be found in Table \ref{tab:tableControllerParametersSim}. 

To make the process of tuning the control gains $K_p$ and $K_d$  less arbitrary and, to some extent, repeatable, optimal gains were identified by using a genetic algorithm. For this automated tuning process, we defined a combined reference trajectory chaining together, an ellipse type and a Lissajous type trajectory. The parameters for both trajectory types were selected to be mean values of the parameter ranges used in the Monte Carlo simulations (see Table \ref{tab:trajectory-parameters})). Based on this reference trajectory, we employed the \emph{geneticalgorithm} package~\cite{genetic2020} to implement the Genetic Algorithm (GA) optimization technique for tuning the gains of the controller. We used the following GA algorithm parameters: a maximum number of iterations of $50$, a population size of $20$, a mutation probability of $0.45$, an elitism ratio of $0.01$, a crossover probability of $0.5$, a parents portion of $0.25$, and a uniform crossover type.

We defined a cost function $\Upsilon$ to be minimized such that:

\begin{equation}
\Upsilon = \sqrt{\frac{1}{N} \sum_{j=1}^{N} \prescript{\Phi}{}\eta_{e, j} Q^{GA} \prescript{\Phi}{}\eta_{e, j} + \tau^{sim}_j R^{GA} \tau^{sim}_j}
\end{equation}

Where $N$ is the total number of iterations, $Q^{GA}$ and $R^{GA}$ are positive-definite weighting matrices such that $Q^{GA}=diag([100,100,100,50,50,50,50])$ and $R^{GA}=diag(0.5,0.5,0.5,0.5,0.5,0.5)$. Finally, the optimization was constrained by predefined boundaries for $K_p$ and $K_d$ of $0.1$ to $50.0$.


\begin{table*}[tbh!]
\centering
\caption{Controller and control allocation parameters used in the simulation.}
\begin{adjustbox}{width=0.65\textwidth}
\small
\begin{tabular}{|c|c|c|c|c|}
\cline{1-4}
   & $CA_{prop}$ & $CA_{opt}$ & $CA_{inv}$  \\ \hline
\multicolumn{1}{|c|}{$K_{p}$} & $[3.81, 3,39, 3.76, 3.93]$ & $[4.25, 3.04, 2.79 4.14]$ & $[17.45, 4.31, 7.41, 14.65]$\\ \hline
\multicolumn{1}{|c|}{$K_{d}$} & $[3.46, 4.59, 4.41, 2.01, 3.39, 4.68]$ & $[2.8, 4.0, 4.35, 1.89, 3.43, 3.7]$ & $[16.02, 48.5, 8.83, 18.35, 6.07, 42.41]$ \\ \hline
\multicolumn{1}{|c|}{$n_d$} & \multicolumn{3}{c|}{$[2, 2, 4, 4, 4, 2]$} \\ \hline
\multicolumn{1}{|c|}{$K_{amp}$} & \multicolumn{2}{c|}{$10$}  & $5$ \\ \hline
\multicolumn{1}{|c|}{$K_{zd}$} & \multicolumn{2}{c|}{$3$} & $2$ \\ \hline
\multicolumn{1}{|c|}{$\alpha_{comp}$} & $30$ & \multicolumn{2}{c|}{-}\\ \hline
\multicolumn{1}{|c|}{$\overline{f^{th}}_{max}$} & \multicolumn{3}{c|}{$5$} \\ \hline
\multicolumn{1}{|c|}{$\Lambda$} & \multicolumn{3}{c|}{$[1.0, 1.0, 1.0, 1.0, 1.0, 1.0, 1.0]$} \\ \hline
\multicolumn{1}{|c|}{$\Gamma$} & \multicolumn{3}{c|}{$[1.0_{23  \times 1}]$} \\ \hline
\multicolumn{1}{|c|}{$\theta_{start}$} & \multicolumn{3}{c|}{$[0_{4 \times 1}, 50.0_{3 \times 1}, 1.0_{3 \times 1}, 0.1, -5.0, -50.0, -10.0, 0_{2 \times 1}, -0.5, -10.0, -100.0, -200.0, -1.0_{2 x 1}, -0.1]$} \\ \hline
\multicolumn{1}{|c|}{$\ubar{\theta}$} & \multicolumn{3}{c|}{$[-2.0, -1.0_{3 \times 1}, 0_{6 \times 1}, -5.0, -10.0, -50.0, -10.0, -5.0_{2 \times 1}, -0.5, -50.0, -500.0_{2 \times 1}, -2.0, -5.0, -1.0]$} \\ \hline
\multicolumn{1}{|c|}{$\bar{\theta}$} & \multicolumn{3}{c|}{$[2.0, 1.0_{3 \times 1}, 100.0_{3 \times 1}, 5.0_{4 \times 1}, 0_{12 \times 1}]$} \\ \hline
\multicolumn{1}{|c|}{$\bar{\epsilon}$} & \multicolumn{3}{c|}{$[0.1_{4 \times 1}, 10.0_{3 \times 1}, 0.5_{3 \times 1}, 0.1,  1.0_{3 \times 1}, 0.1_{3 \times 1}, 5.0, 10.0_{2 \times 1}, 0.1_{3 \times 1}]$} \\ \hline
\multicolumn{1}{|c|}{$\varsigma$} & \multicolumn{3}{c|}{$0.1$} \\ \hline

\end{tabular}
\end{adjustbox}
\label{tab:tableControllerParametersSim}
\end{table*}

We tuned the parameters for the control allocation modules as follows: The CPG parameters $K_{amp}$, $K_{zd}$ and $\alpha_{comp}$ were tuned by running simulations with gains in range $[1-15]$, $[1-15]$, $[0-150]$ respectively, and assessing the allocation errors ${AE_{lin}}$ and ${MAE_{ang}}$ for the three studied control allocation methods. The CPG parameters $K_{amp}$, $K_{zd}$ for $CA_{inv}$ resulted in lower values. This can be explained as, first, to decrease the rotational speed when moving from one zero-direction angle to another, and second, to minimize the oscillating amplitude while performing the zero-direction change. Studying the effect of $\alpha$ was relevant only for the case of the proposed control allocation method CA$_{prop}$. The numeric values for the parameters found are summarized in Table \ref{tab:tableControllerParametersSim}. Moreover, the parametric values for the force to amplitude model denoted in \eqref{eq:force2amplitude} are summarized in Table \ref{tab:tableConstants}.

\begin{table}[t]
\centering
\caption{Parameters for simulating fin dynamics and force to amplitude conversion.}
\label{tab:tableConstants}
\begin{tabular}{
    |p{1.2cm}| 
    p{0.9cm}| 
    p{1.2cm}| 
    p{0.8cm}|
    p{0.7cm}|
    p{0.5cm}|
    p{0.5cm}|
}
\hline
\textbf{$\rho$ $(kg/m^3)$} & \textbf{$S_f$ $(m^2)$} & \textbf{$\omega^{osc}$ ($\frac{rad}{s})$} & \textbf{$r_c$ $(m)$} & \textbf{$C_d$} & \textbf{$C_{Lmax}$} & \textbf{$C_{Dmax}$} \\ \hline
$997$ & $0.02$ & $4\pi$ &$0.1$ & $0.24$ & $1.65$ & $3.2$ \\ \hline
\end{tabular}
\end{table}

Finally, the covariance matrices $R^{EKF}$ and $Q^{EKF}$ for the EKF were hand-tuned based on data from simulations and preliminary experiments. 

\section{Experiments}
\label{sec:experiments}

\subsection{Experimental Setup}
A series of validation experiments were conducted in a swimming pool (see the supplementary video for details). The experimental setup included a fabric grid of size \SI{3}{m} by \SI{6}{m} containing $324$ ArUco markers \cite{garrido2014automatic}. The grid consisted of $108$ markers of size \SI{0.25}{m} and $216$ markers of size \SI{0.1}{m}. These markers were used to provide the robot with position measurements in the Earth-fixed frame $R_n$ using the onboard Chameleon PointGrey camera. The robot's EKF is updated with position measurements from the detected ArUco markers with a frequency of \SI{10}{Hz} and with angular states from the onboard MPU-6050 IMU with a frequency of \SI{100}{Hz}. The tether was manually managed in the pool to limit the disturbances.

Two different control allocation methods were tested : the proposed analytic control allocation ($CA_{prop}$) and the state-of-the-art optimization-based control allocation ($CA_{opt}$). The SQP required for the optimization is provided by the library \emph{ALGLIB}~\cite{ALGLIB}. The naive control allocation method ($CA_{inv}$) was excluded from these experimental trials due to its demonstrated ineffectiveness in simulations~\ref{subsec:control_allocation_comparison}.

The experiments were performed for the 3-degrees-of-freedom tracking scenario with roll and pitch stabilized (3T2S), employing two different trajectory types: ellipse and Lissajous. For each trajectory type and for each of the two tested control allocation methods, five trials were conducted, resulting in a total of 20 trials overall.

\color{black}
The full 6T scenario was not considered in these experiments due to several limitations related to the vehicle's design and practical constraints in the experimental setup, particularly, the potential loss of ArUco marker-based position feedback when controlling roll and pitch. Additionally, the distribution of centre of mass and centre of buoyancy created a passively stabilizing system which would have disturbed the controller and potentially exceeded the available control authority. Furthermore, in its current setup, the controller admits fin-induced roll and pitch motions via the state estimation directly into the control cycle, which leads to increased chattering and controller band-width saturation. 

To demonstrate the capability of our full control allocation method in a real-world scenario despite the limitations, we conducted a 5-DOF motion control experiment, referred to as the 3T2F scenario, employing $CA_{prop}$ control allocation approach. In this experiment, the robot was tasked with following a straight-line trajectory in three-dimensional space with desired surge and heave velocities of $\dot{x}_d = \SI{0.075}{m/s}$ and $\dot{z}_d = \SI{0.0075}{m/s}$ respectively, and a surge offset of \SI{0.3}{m}. To mitigate the control challenges previously discussed, we applied constant feedforward force setpoints of \SI{0.25}{Nm} to both the roll and pitch degrees of freedom. While achieving full trajectory tracking in 5-DOF remains a goal for future work, requiring improvements in the robot's design and control algorithms, as highlighted in the discussion of the tracking results, we demonstrate control allocation across five DOF through this experiment.

\color{black}
Many of the metrics used in simulation could not be directly transferred to the evaluation of the experimental results as essential parameters such as forces produced by the fins could not be measured. Therefore, the tracking errors for each DOF are presented as well as a root mean square metric for demanded forces/torques for every DOF. Finally, we provide mean and maximum computation times for the control allocation. 


\begin{table}
\centering
\caption{Numeric values of trajectory parameters used in experiments}
\label{tab:trajectory-parameters-exp}
\tiny
\begin{tabular}{c|c|c|c|c|c|c|c|c|c|c|}
\cline{2-11}
& \multicolumn{1}{c|}{$x_{0}$} & \multicolumn{1}{c|}{$A_{x}$} & \multicolumn{1}{c|}{$\omega_x$} & \multicolumn{1}{c|}{$A_{y}$} & \multicolumn{1}{c|}{$\omega_y$} & \multicolumn{1}{c|}{$A_{z}$} & \multicolumn{1}{c|}{$\omega_z$} & \multicolumn{1}{c|}{$t*$} & \multicolumn{1}{c|}{$\gamma_1$} & \multicolumn{1}{c|}{$\gamma_2$} \\ \hline
\multicolumn{1}{|c|}{Ellipsoidal} & \multicolumn{1}{c|}{0.3} & \multicolumn{1}{c|}{1.75} & \multicolumn{1}{c|}{0.06} & \multicolumn{1}{c|}{1.0} & \multicolumn{1}{c|}{0.06} & \multicolumn{1}{c|}{0.2} & \multicolumn{1}{c|}{0.03} & \multicolumn{1}{c|}{0.2} & \multicolumn{1}{c|}{7.0} & \multicolumn{1}{c|}{1.0} \\ \hline
\multicolumn{1}{|c|}{Lissajous} & \multicolumn{1}{c|}{0.3} & \multicolumn{1}{c|}{1.5} & \multicolumn{1}{c|}{0.045} & \multicolumn{1}{c|}{0.7} & \multicolumn{1}{c|}{0.045} & \multicolumn{1}{c|}{0.25} & \multicolumn{1}{c|}{0.03} & \multicolumn{1}{c|}{0.2} & \multicolumn{1}{c|}{7.0} & \multicolumn{1}{c|}{1.0} \\ \hline
\end{tabular}
\end{table}

\subsection{Hyper-parameter tuning in experiments}

Starting from the identified values in the simulation, the controller parameters $K_p$ and $K_d$ were adjusted using a manual tuning process based on pool trials with $CA_{prop}$. The same parameters were then tested directly with the optimization-based control allocation method. Observations indicated that the values required no additional adjustments and provided a satisfactory performance. The controller parameters for both control allocation methods were therefore defined as $K_p = [1.9, 0.1, 3.7, 1.6]$ and $K_d = [8.5, 0.5, 8.8, 1.16, 2.58, 3.5]$. For the additional 5-DOF experiment we used the following controller gains: $K_p = [1.2, 0.1, 3.8, 1.2]$ and $K_d = [8.5, 0.5, 19.8, 1.16, 5.0, 6.8]$


\begin{figure}[t]
    \centering
    \includegraphics[width=\linewidth]{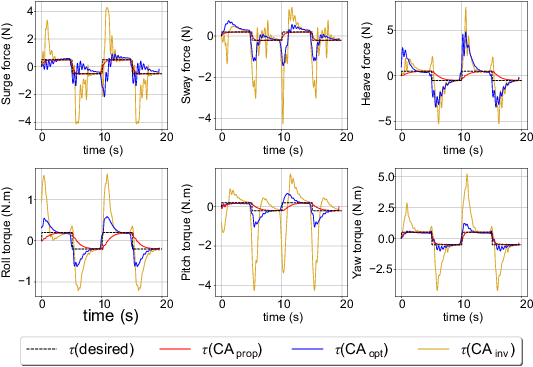}
    \caption{Comparison of the studied control allocation approaches for tracking a desired control input.}
    \captionsetup{justification=centering,margin=2cm}
    \label{fig:compareCA}
\end{figure}


The parameters for adaptive control, control allocation, trajectory generation and EKF obtained during the simulation tuning process (summarized in T\ref{tab:tableControllerParametersSim}) were kept the same in experiments.

\section{Results and discussion}


\subsection{Control allocation comparative results}
\label{subsec:control_allocation_comparison}
The performance of the investigated control allocation methods was first evaluated in the simulation environment in isolation by establishing a desired 6-DOF wrench vector, $\tau^{des} = [0.5, 0.5, 0.5, 0.2, 0.2, 0.2]^T$, programmed to switch sign every 5 seconds, as depicted in Fig. \ref{fig:compareCA}. The mean wrenches in body frame produced by the fins under the influence of the various control allocation methods are displayed in Figure \ref{fig:compareCA}.

As displayed in Fig. \ref{fig:compareCA}, CA$_{prop}$ generates outputs that align with the desired wrenches $\tau^{des}$ without any overshoot. This can be attributed to the minimization of the zero-direction change implemented in this method. A slight response delay can be noted for heave, roll, and pitch. In terms of $MAE_{lin}$ and $MAE_{ang}$, as presented in Table \ref{tab:comparison_CA_methods}, the proposed method CA$_{prop}$ yields the most effective results.

In contrast, CA$_{opt}$ exhibits a rapid response in the generated body forces and torques, but with a significant overshoot. This is mainly due to the lateral fin forces that emerge when the fins' zero-direction angles change abruptly, which results in greater linear $MAE_{lin}$ and angular $MAE_{ang}$ allocation errors, compared to the proposed control allocation method CA$_{prop}$.

Regarding CA$_{inv}$, the obtained results indicate that this method is not well-suited for fin-actuated vehicle control. This is because all fins contribute to wrenches in all DOF, without considering the change in the zero-direction angle. Fast or large changes in zero direction then cause a substantial overshoot in body forces, as evidenced in Figure \ref{fig:compareCA}.

These findings highlight the superior performance of the proposed method CA$_{prop}$ over the other two control allocation methods in terms of force allocation errors. This superior performance is due to the fact that CA$_{prop}$ is analytically designed to minimize the change in zero-direction angle. For CA$_{opt}$, even though the change in zero-direction angle is not considered in the optimization cost function, the optimization algorithm tends to favor the solution closest to the initial conditions of the fins' zero-directions. This results in notably better performance compared to the pseudo-inverse method CA$_{inv}$.

\begin{table}[h]
		\centering
		\vspace{1.5mm}
		\setlength\doublerulesep{0.5pt}
		\caption{Comparison of control allocation methods (CA$_{prop}$, CA$_{opt}$, and CA$_{inv}$) in terms of mean allocation errors for linear ($MAE_{lin}$) and angular ($MAE_{ang}$) DOF.}
		{\renewcommand{\arraystretch}{1.2}
		\begin{tabular}{l|l|l|l|}
		 & \textbf{CA$_{prop}$} & \textbf{CA$_{opt}$} & \textbf{CA$_{inv}$}  \\ \hhline{=|=|=|=}
	       $MAE_{lin}$ [N] & \textbf{0.293} & 1.369 & 2.451 \\
              $MAE_{ang}$ [Nm] & \textbf{{0.206}} & 0.467  & 2.613
        \end{tabular}}
		\label{tab:comparison_CA_methods}
\end{table}



\subsection{Trajectory tracking simulation}
Tables \ref{tab:summaryStatistics1} and \ref{tab:summaryStatistics2} summarize the results for the 6T and 3T2S scenarios respectively. For the majority of evaluation metrics, the proposed method (CA$_{prop}$) outperformed the two other methods. It achieved the lowest RMSE in both linear and angular measures, implying better tracking performance. The difference in tracking performance between CA$_{prop}$ and CA$_{inv}$ is very notable, going from small errors of \SI{4.3}{cm} and \SI{3.4}{\degree} to significant deviations of \SI{1.15}{m} and \SI{103}{\degree}. The simple pseudo-inverse based control allocation is clearly inadequate to provide satisfactory tracking. CA$_{inv}$ creates too much disturbance during fin rotations, because all fins are considered for providing propulsion in each DOF. This is minimized by explicitly choosing the number of fins to contribute to each DOF and the introduced force compensation in CA$_{prop}$, leading to significantly better tracking. In CA$_{opt}$, the minimization of fin rotations is not explicitly defined, but seems to be implicitly taken into account. 

Furthermore, CA$_{prop}$ demonstrated the smallest median actuation efforts (MAW, MAE$_{lin}$, MAE$_{ang}$), which suggests more efficient energy usage compared to the other tested methods. Again, the differences to CA$_{inv}$ are significant, whereas the differences to CA$_{opt}$ are close or within the variability of the reported metrics.

Furthermore, CA$_{prop}$ and CA$_{opt}$ demanded significantly less forces with better tracking results compared to CA$_{inv}$. CA$_{prop}$ had the smallest MAW while CA$_{opt}$ had the smallest MW, although the differences are very small in both cases. CA$_{inv}$, however, had maximum wrench demands that were well outside of the maximum wrench magnitude of \SI{12.5}{N} indicating that actuators were demanded to operate above their limits. 

In terms of computation time for control allocation CA$_{inv}$ is the fastest, which can be expected as it includes the least amount of computations. Even though not the fastest, CA$_{prop}$'s MCT is significantly less than the CA$_{opt}$, highlighting the main advantage of our proposed approach over the optimization based solution. Especially for resource constraint systems, the roughly 40 fold decrease in computation time can be very relevant. Additionally, at high update rates, a median MCT of \SI{3.33}{ms} can create a computational bottleneck for systems using CA$_{opt}$ when other computationally resource heavy algorithms for navigation and planning are to be employed too. Moreover, the much higher MCT also leads to a significantly higher effort to compute the controller gains with the GA significantly increasing run-time and energy consumption. 



\begin{table}[h]
		\centering
		\vspace{1.5mm}
		\setlength\doublerulesep{0.5pt}
		\caption{Summary statistics for represented by median (IQR), of defined evaluation metrics for MonteCarlo simulation framework with 500 trials in the 6T scenario. Results are presented for 3 different control allocation scenarios: CA$_{prop}$, CA$_{opt}$, CA$_{inv}$.}
		{\renewcommand{\arraystretch}{1.2}
		\begin{tabular}{l|l|l|l|}
		\textbf{Summary Statistic} & \textbf{CA$_{prop}$} & \textbf{CA$_{opt}$} & \textbf{CA$_{inv}$}  \\ \hhline{=|=|=|=}
			$RMSE_{lin}$ [m] & \textbf{\textcolor{black}{0.04 (0.02)}} & 0.05 (0.01) & 1.15 (0.3)\\
			$RMSE_{ang}$ [rad] & \textbf{\textcolor{black}{0.06 (0.003)}} & 0.07 (0.008) & 1.83 (0.18) \\
            $MEM_{lin}$ [m] & \textbf{\textcolor{black}{0.15 (0.01)}} & \textbf{\textcolor{black}{0.15 (0.01)}} & 2.4 (0.72) \\
            $MEM_{ang}$ [rad] &\textbf{\textcolor{black}{0.53 (0.04)}} & 0.55 (0.06) & 4.42 (0.48) \\
            $MAW$ [N] & \textbf{\textcolor{black}{1.34 (0.28)}} & 1.69 (0.17) & 6.93 (1.02) \\
            $MW$ [N] & \textbf{\textcolor{black}{4.19 (0.22)}} & 4.34 (0.28) & 38.68 (6.52) \\
            $MCT$ [ms] & 0.08 (0.002) & 3.3 (0.19) & \textbf{\textcolor{black}{0.07 (0.002)}} \\
            $MAE_{lin}$ [N] & \textbf{\textcolor{black}{0.48 (0.04)}} & 0.71 (0.03) & 5.0 (0.34) \\
            $MAE_{ang}$ [Nm] & \textbf{\textcolor{black}{0.32 (0.03)}} & 0.42 (0.03) & 1.37 (0.11)
        \end{tabular}}  
		\label{tab:summaryStatistics1}
\end{table}

The metrics show no discernible difference between the two tracking scenarios, indicating that the trajectory based sway compensation was successful in compensating the lack of sway forces. The results for both tracking scenarios suggest that the proposed CA$_{prop}$ control allocation method offers a compelling balance between high performance and actuator efficiency, with considerable advantages in computational efficiency compared to CA$_{opt}$. This makes CA$_{prop}$ a promising solution for both 6T and 3T2S scenarios for fin-actuated underwater robots.  

\begin{table}[h]
		\centering
		\vspace{1.5mm}
		\setlength\doublerulesep{0.5pt}
		\caption{Summary statistics, represented by median (IQR), of defined evaluation metrics for Monte Carlo simulation framework with 500 trials in the 3T2S scenario. Results are presented for 3 different control allocation scenarios: CA$_{prop}$, CA$_{opt}$, CA$_{inv}$.}
		{\renewcommand{\arraystretch}{1.2}
		\begin{tabular}{l|l|l|l|}
		\textbf{Summary Statistic} & \textbf{CA$_{prop}$} & \textbf{CA$_{opt}$} & \textbf{CA$_{inv}$}  \\ \hhline{=|=|=|=}
			$RMSE_{lin}$ [m] & \textbf{\textcolor{black}{0.04 (0.02)}} & 0.05 (0.02) & 2.1 (1.21)\\
			$RMSE_{ang}$ [rad] & \textbf{\textcolor{black}{0.06 (0.02)}} & 0.09 (0.05) & 1.74 (0.12) \\
            $MEM_{lin}$ [m] & \textbf{\textcolor{black}{0.13 (0.03)}} & 0.15 (0.05) & 4.17 (2.14) \\
            $MEM_{ang}$ [rad] & \textbf{\textcolor{black}{0.34 (0.49)}} & 0.7 (0.67) & 3.22 (0.07) \\
            $MAW$ [N] & \textbf{\textcolor{black}{1.27 (0.24)}} & 1.6 (0.15) & 4.87 (2.07) \\
            $MW$ [N] & 4.09 (0.12) & \textbf{\textcolor{black}{3.86 (0.34)}} & 34.35 (6.87) \\
            $MCT$ [ms] & 0.08 (0.009) & 3.8 (0.55) & \textbf{\textcolor{black}{0.07 (0.002)}} \\
            $MAE_{lin}$ [N] & \textbf{\textcolor{black}{0.42 (0.02)}} & 0.63 (0.06) & 4.66 (1.12) \\
            $MAE_{ang}$ [Nm] & \textbf{\textcolor{black}{0.34 (0.03)}} & 0.47 (0.04) & 1.37 (0.15)
        \end{tabular}}
		\label{tab:summaryStatistics2}
\end{table}

\color{black}
\subsection{Experimental Results - 3T2S Scenario}
\color{black}

\begin{figure}[h]
    \centering
    \includegraphics[width=0.8\linewidth]{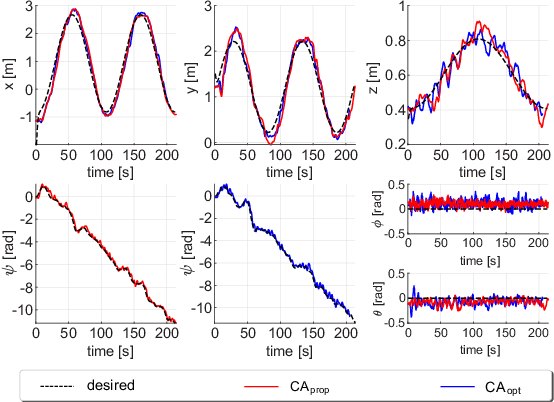}
    \caption{Experimental tracking results for the 3D ellipsoidal trajectory: In dotted black line is the desired trajectory. In red line is performed trajectory using the proposed solution CA$_{prop}$. In blue line is performed trajectory using the optimal solution CA$_{opt}$.}
    \captionsetup{justification=centering,margin=2cm}
    \label{fig:trackingHelix}
\end{figure}

\begin{figure}
    \centering
    \includegraphics[width=0.8\linewidth]{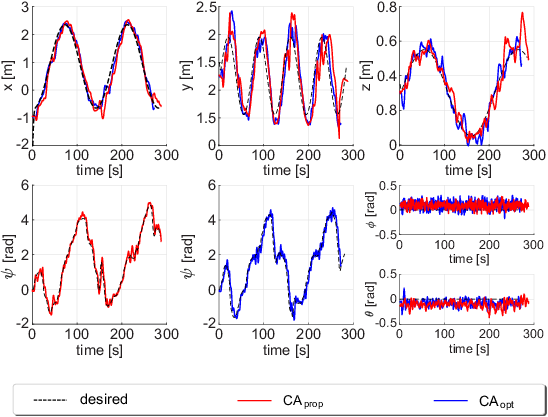}
    \caption{Experimental tracking results for the 3D Lissajous trajectory: In dotted black line is the desired trajectory. In red line is performed trajectory using CA$_{prop}$ solution. In blue line is performed trajectory using CA$_{opt}$ solution.}
    \captionsetup{justification=centering,margin=2cm}
    \label{fig:trackingLissajous}
\end{figure}

Fig.~\ref{fig:trackingHelix} and Fig.\ref{fig:trackingLissajous} respectively show the ellipse and Lissajous trajectory tracking results of an example trial. The results demonstrate that both the proposed and optimal control allocation solutions provide good tracking performance for all controlled DOF. Moreover, the line of sight implementation allowed for the tracking of the non-directly commanded sway component of the trajectory. While the roll and pitch components were not actively tracked, the robot's passive stability in these DOF maintained their tracking error close to zero with slight oscillations due to the oscillatory actuation method. Quantitatively, the results indicate that the optimal solution CA$_{opt}$ slightly outperformed the proposed solution CA$_{prop}$ in terms of tracking accuracy. The largest median error was observed in tracking the surge component for the Lissajous trajectory, with CA$_{prop}$ having a median error of \SI{41}{cm} and CA$_{opt}$ having a median error of \SI{25}{cm}. The Lissajous trajectory was slightly more complex with more orientation changes, so a slight decrease in tracking performance can be expected. However, the results do show that the control framework does provide adequate performance for various trajectories. For both allocation methods the surge DOF exhibited the biggest errors, which in part could be attributed to the choice of controller gains, which were set to prevent the robot from overtaking the desired trajectory $x_d$, which would have resulted in the robot making a full turn.


Generally, for both trajectories the tracking errors in the experiments were significantly higher compared to the simulation results. This can be explained by several simplifications and assumptions made for the simulations. During simulations we did not simulate the low-level motor control and assumed perfect and instantaneous tracking of the fin kinematics demanded by the CPG. However, for the physical vehicle , amplitude and frequency can not be perfectly tracked by the fin motors and each fin will exhibit certain time-lags. Additionally, the employed fin model \eqref{eq:FinLiftDrag} does not capture complex fluid-body interactions between the flexible fins and the surrounding water or between vortices created by  fins facing each other. In simulations we also did not assume any environmental disturbances. The experiments, however, were conducted in a public bath during regular opening hours.  This is also a potential explanation for CA$_{opt}$ slightly outperforming CA$_{prop}$, which is in contrast to the simulation results. During tests with CA$_{opt}$ the amount of people in other  lanes of the swimming pool and subsequent disturbances were less compared to tests with CA$_{prop}$. Given the small absolute differences between the algorithms \SI{10}{cm} to \SI{15}{cm} and below \SI{1}{\degree} such a small environmental effects can have a potential effect in the results. Nevertheless, it can be concluded that the optimization based approach slightly outperformed the proposed allocation.

When comparing our RMSEs of maximal \SI{8.5}{cm} in depth tracking and \SI{11}{\degree} in yaw tracking to the errors reported by previous control attempts for U-CAT or similar vehicles, \SI{2.22}{cm} in depth set point stabilization and \SI{2.91}{\degree} in yaw tracking \cite{salumae2016motion}, \SI{2.28}{cm} \cite{chemori2016depth} in depth tracking and \SI{1.2}{cm} $\pm$ \SI{5.3}{cm} again for depth tracking only, our depth and yaw tracking is slightly less accurate. However, in \cite{chemori2016depth} and \cite{giguere2013wide} only regulation to a constant depth without any tracking in other DOF was considered, while in \cite{salumae2016motion} depth was again only regulated to a set-point and yaw was effectively decoupled from surge and heave. Taking this into account we show here a similar performance, while tracking more DOF than before. This comparison shows that the agility and full potential of fin driven turtle like AUVs can be utilized by expanding the tracked DOF while maintaining a similar level of accuracy compared to the state-of-the-art. \\

Given that we implemented the hybrid adaptive controller from Basso et al. \cite{basso2022global}, it makes sense to compare to their experimental results as well. A direct comparison was not possible due to the lack of numerical results. However, based on visual observation of the presented results it seems our tracking performance lies in a similar range, albeit slightly less accurate. We believe that this shows reproducibility of the results from \cite{basso2022global} in a different and complex setting. When comparing the setup to Basso et al. \cite{basso2022global} it should be noted that we can show a similar tracking performance with half the amount of actuators (8 vs 4) on a functionally non-holonomic system that lacks sway actuation, which highlights the applicability of the hybrid adaptive controller to additional complex robotic platforms, highlighting the efficacy of the control framework proposed by Basso et al. Additionally, we used a very rough approximation of the identified dynamics model \cite{salumae2017motion} to populate the starting values for the adaptive parameters, whereas in \cite{basso2022global} the starting values for the adaptive parameters where directly derived from an identified dynamics model. It should be noted however, that the low velocity trajectories in both our experiments and in \cite{basso2022global} lead to limited excitations, which puts more weight on the starting parameters and supports the choice to start with an identified model in \cite{basso2022global}

We conclude, that our results show a satisfactory performance compared to the state-of-the-art in terms of control. Furthermore, we see a significant contribution of our work in the independent and successful replication of the work presented in \cite{basso2022global} on a more complex system in terms of actuation. Additionally, we extended the applicability of the controller proposed by Basso et al. to non-holonomic systems. \\ 



\begin{figure}[h]
    \centering
    \includegraphics[width=0.8\linewidth]{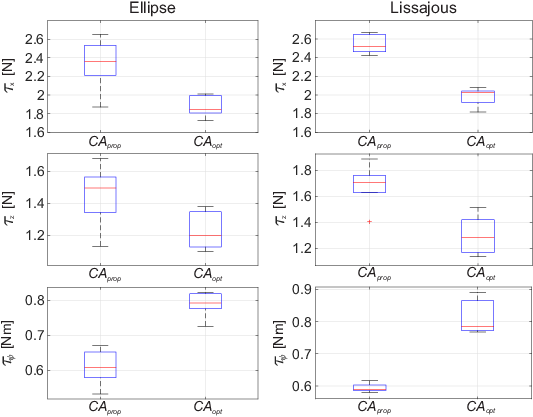}
    \caption{Box-plot of root mean square forces and torques. Left: During ellipse trajectory. Right: During Lissajous trajectory.}
    \label{fig:forcesBoxPlot}
\end{figure}

\begin{figure}[h]
    \centering
    \includegraphics[width=1.0\linewidth]{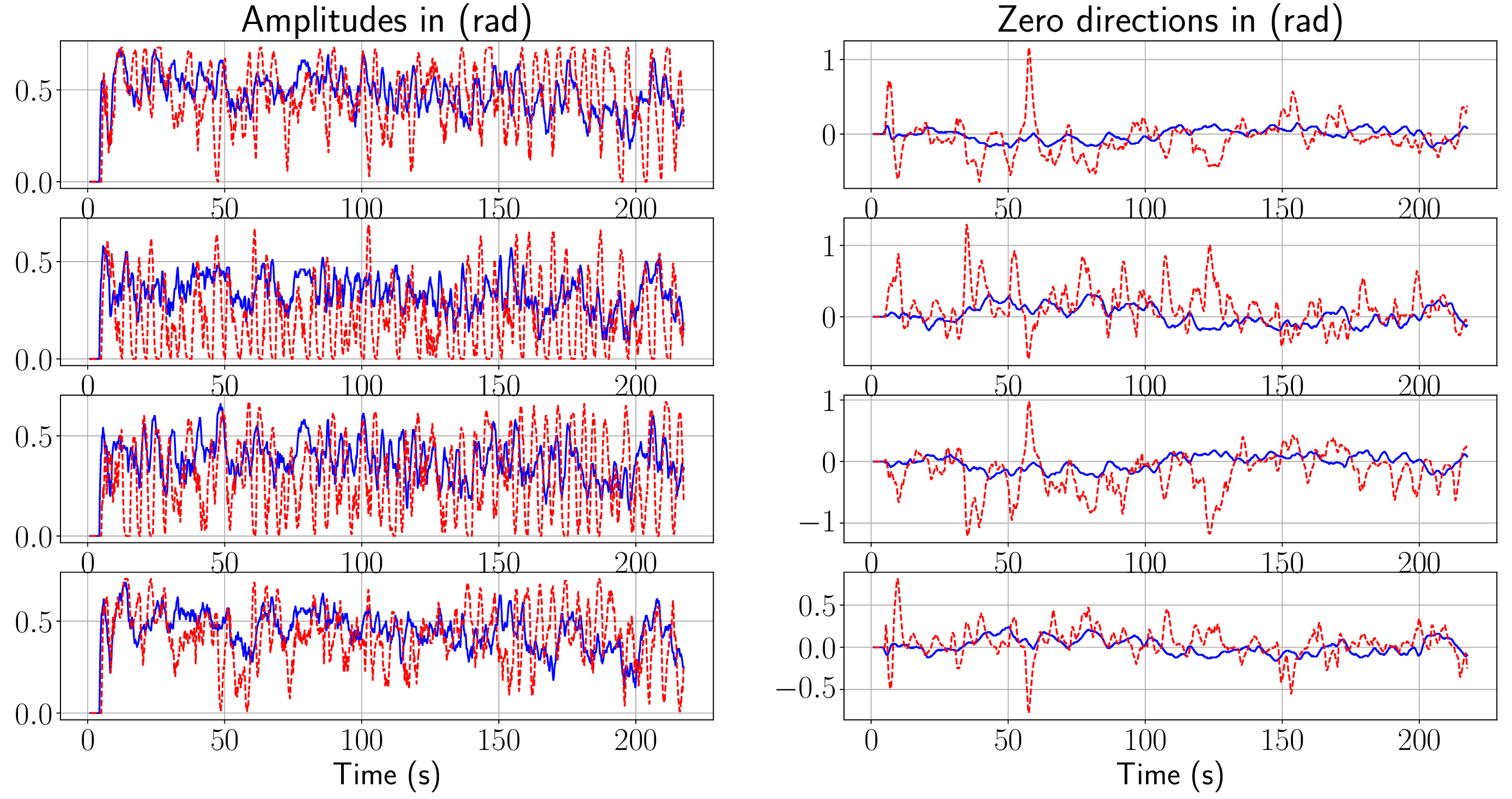}
    \caption{Comparison of fin kinematic oscillation parameters  ($\phi_{0, i}$) for $i = 1 \hdots 4$ from top to bottom: Each subplot compares the results from the two different allocation methods: \textit{CA\textsubscript{prop}} (blue solid line) and \textit{CA\textsubscript{opt}} (red dashed line).}
    \label{fig:finkinematics}
\end{figure}

Fig.~\ref{fig:forcesBoxPlot} shows the box-plot of the root mean square (RMS) of the generated controller outputs ($\tau^{des})$ for both control allocation methods. The results indicate that the controller generated slightly higher forces in the linear directions surge and heave for CA$_{prop}$, and slightly lower torque for yaw. This can be explained by the introduction of the force compensation term $f_{comp}$ \eqref{eq:fcomp} that minimizes change in zero-direction angle at the cost of an increase in necessary forces to compensate for the change in the thrust vector.  Moreover, Fig.~\ref{fig:finkinematics} shows a qualitative comparison between two trials for the same ellipsoidal trajectory using the two studied allocation methods. The proposed method CA$_{prop}$ offers much smoother and smaller zero direction angles, resulting in minimal disturbances by the lateral fin forces. In contrast, the optimization-based method results in large variations in the generated zero directions. The amplitudes for CA$_{prop}$ show that the distributed forces remain almost equally distributed among the fins, whereas for CA$_{opt}$, the generated amplitudes have slightly higher peaks, but are also almost non-actuated with amplitudes close to zero in several short time intervals. This confirms a trade-off between disturbance minimization and energy efficiency regarding the actuators, which needs to be considered for each specific application.

\begin{figure}[h]
    \centering
    \includegraphics[width=\linewidth]{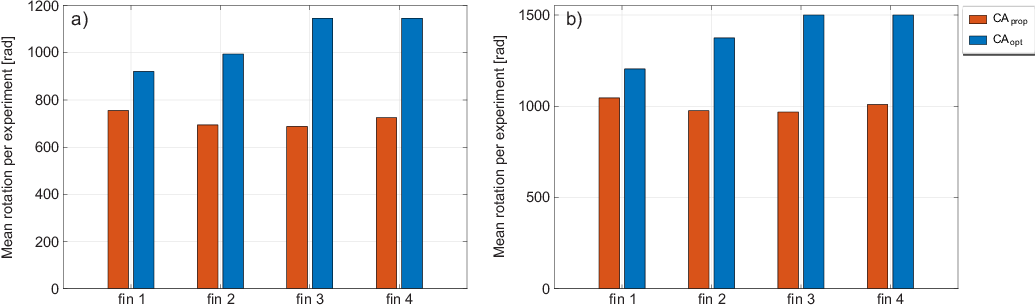}
    \caption{Mean accumulated rotation of U-CAT's four fins per experiment for a) the ellipse trajectory and b) the Lissajous trajectory.}
    \label{fig:barPlotfinRotationCombined}
\end{figure}

The computation times for the two control allocation methods were calculated on the robot's Jetson TX2 embedded computer and are compared in Table \ref{tab:comp_times}.  For the CA$_{prop}$ method, the mean computation times for both ellipse and Lissajous trajectories were very similar, approximately \SI{0.0068}{ms} and \SI{0.0067}{ms} respectively. The reported maximum computation time for both trajectories was less than a third of a millisecond. In contrast, the CA$_{opt}$ method needed considerably longer computation times. The mean times were about \SI{22}{ms} and \SI{21}{ms} for the ellipse and Lissajous trajectories respectively. The maximum times increased drastically for the CA$_{opt}$ method, with $\approx$ \SI{3931}{ms} for the ellipse and $\approx$ \SI{1812}{ms} for the Lissajous trajectory. The difference in computation times is higher than in the simulations, given that the the Jetson TX2 as embedded computer has less computational capacity compared to a laptop. This highlights the big advantage of our proposed control allocation method. Finally, Fig.~\ref{fig:barPlotfinRotationCombined} a) and Fig.~\ref{fig:barPlotfinRotationCombined} b) show the mean accumulated rotations by the four fins for the ellipse and Lissajous trajectories, respectively. The results indicate that CA$_{prop}$ demands less in terms of actuation efforts, which results in better energy efficiency. \\
Taking all relevant metrics into account one can state that both $CA_{opt}$ and $CA_{prop}$ provide a satisfactory tracking performance with $CA_{opt}$ performing better in terms of tracking, while $CA_{prop}$ is slightly more efficient in terms of energy usage and significantly more efficient in terms of computation. With this trade-off it seems the choice of algorithm is depending on the respective application and its demands regarding accuracy and efficiency. However, we believe that $CA_{prop}$ offers significant benefits specifically for resource constraint system, especially if additional computationally complex algorithms for mapping, localization or trajectory generation should run on the same system. \\



\begin{table}[]
\centering
\caption{Computation times for CA$_{prop}$ and CA$_{opt}$ methods}
\label{tab:comp_times}
\begin{tabular}{c|cc|cc|}
\cline{2-5}
                                                 & \multicolumn{2}{c|}{CA$_{prop}$}        & \multicolumn{2}{c|}{CA$_{opt}$}            \\ \hline
\multicolumn{1}{|c|}{Trajectory type}            & \multicolumn{1}{c|}{Ellipse}  & Lissajous & \multicolumn{1}{c|}{Ellipse}     & Lissajous \\ \hline
\multicolumn{1}{|c|}{Mean time (ms)} & \multicolumn{1}{c|}{\textbf{\textcolor{black}{0.0068}}} & \textbf{\textcolor{black}{0.0067}}    & \multicolumn{1}{c|}{22.2779}   & 21.0504   \\ \hline
\multicolumn{1}{|c|}{Max time (ms)}  & \multicolumn{1}{c|}{\textbf{\textcolor{black}{0.3199}}} & \textbf{\textcolor{black}{0.1588}}    & \multicolumn{1}{c|}{3931.6030} & 1812.4072 \\ \hline
\end{tabular}
\end{table}

\color{black}
\subsection{Experimental Results - 3T2F Scenario}
\color{black}

\begin{figure}[h!]
    \centering
    \includegraphics[width=\linewidth]{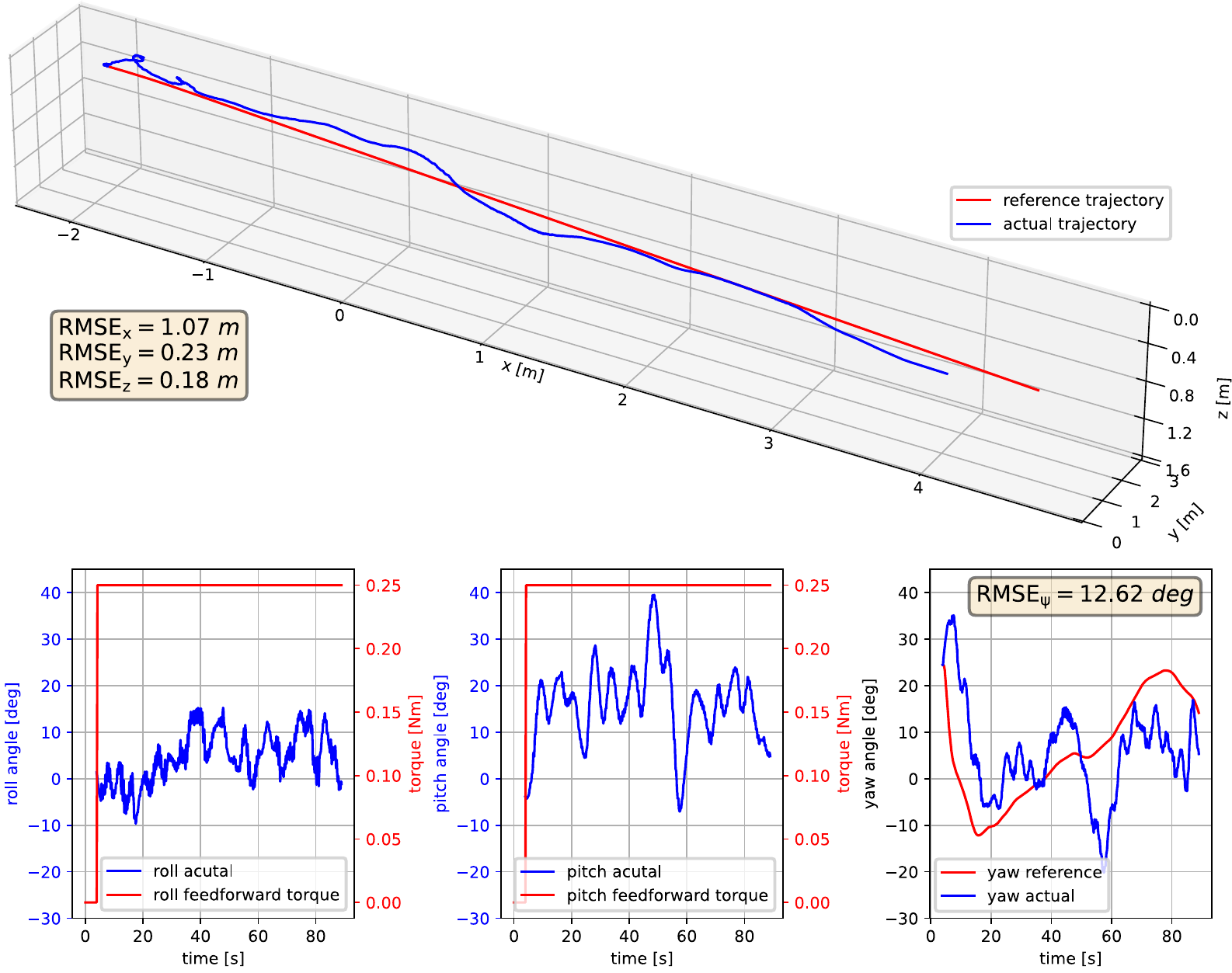}
    \caption{Results for 5-DOF control. a) 3D plot of reference and actual trajectory with RMSE for each linear DOF. b) Results for angular DOF, including force setpoints for roll and pitch with corresponding output and yaw trajectory tracking. }
    \captionsetup{justification=centering,margin=2cm}
    \label{fig:TrajTracking5DOF}
\end{figure}

Finally, results are presented to showcase the control allocation performance of $CA_{prop}$ for all DOF except sway. Fig. \ref{fig:TrajTracking5DOF} shows the tracking results for surge, sway, heave and yaw with corresponding RMSEs as well as the roll and pitch angles due to the constant force set-points. In comparison to the 3T2S setup it can be seen that the RMSEs in sway, heave and yaw are quite similar, whereas tracking in surge is significantly worse. Additionally, from Fig \ref{fig:TrajTracking5DOF} b) it can be seen that the passive restoring forces induce oscillations in the robot even though the set-points for roll and pitch were static. Those oscillations together with the fin induced periodic roll and pitch motions due to fin oscillation and/or rotation create noise in the control system that potentially demands more control action in heave than necessary and reduce the available control action for surge, leading to a reduced performance. However, the results do show that even with an unfavorable ballasting and self induced controller disturbances, the control allocation manages to demand forces in all active DOF.

\color{black}
A significant limitation of our experimental setup was the inability to effectively test roll and pitch tracking. The current robot's configuration, with its center of mass and center of buoyancy creating passive restoring moments, was designed to passively stabilize pitch and roll. However, this passive stabilization, combined with unaccounted internal and periodic disturbances due to fin oscillations, interfered with the ability for active stabilization. These disturbances possibly exceeded the available control authority of the actuators, making active control of roll and pitch challenging. Moreover, balancing the generated forces in fin-actuated systems is inherently more complex than in thruster-based AUVs. With U-CAT's fin actuation, we assume that two fins oscillating with the same parameters generate the same amount of thrust. However, this is rarely the case due to several factors such as variations in motor performance lead to poor tracking of the desired oscillation parameters, resulting in thrust differences between fins. Additionally, hydrodynamic interactions between fins were not taken into account, for instance fins oscillating in opposition can interfere with each other's flow fields, affecting thrust generation and introducing additional disturbances.

These issues are inherent to U-CAT's design but represent important lessons learned. Addressing these challenges requires improvements in motor control precision, more accurate thrust modeling that accounts for flow conditions, and consideration of fin interactions in the control strategy. Another limitation of the work presented here lies within the tested velocities for tracking. Maximum tested velocities approached \SI{0.2}{m/s}, which can be sufficient for many intended monitoring tasks. However, in the presence of strong water currents this limit can constitute a problem. One reason for the limited velocities tested was the available experimental setup, which did not allow the motion over bigger areas, implicitly limiting achievable velocities. Another limitation is the thrust output that can be generated by the current fin design. However, those limitations are mainly the results of engineering challenges and can be overcome. We don't see any fundamental limitations of the presented work in regard to attainable velocities within certain bounds.
\color{black}

Notwithstanding the present limitations we constitute that the obtained simulation and experimental results present a comprehensive analysis of the proposed solution, highlighting its effectiveness and potential for further improvement. The experiments demonstrate that the proposed solution yields good tracking performance for all controlled DOF. Specifically, the methods are shown to effectively track complex 3D trajectories with small tracking error while maintaining efficient control, actuation efforts, and minimal computational time.


\section{Conclusion and future work}
\label{sec:conclusion}

This study presented an in-depth investigation into 6-DOF tracking control using an under- and fin-actuated AUV. The core of our work was the development of an innovative analytic control allocation method for fin-driven AUVs. The allocation is supplemented by the adaptation and extention of a state-of-the-art hybrid adaptive controller, which facilitates globally stable 6-DOF trajectory tracking control. The novel control allocation method allows for simultaneous control of all six DOF using only four actuators. By addressing the specific challenges posed by fin-based actuation, our proposed method achieves computational efficiency compared to state of the art optimization based control allocation, while achieving similar energetic and tracking efficiency.

The developed hybrid controller exhibits significant potential for enabling simultaneous 6-DOF control for an under-actuated fin-actuated AUV, thereby advancing the field towards more sophisticated and versatile underwater robots. This method was tested and validated through extensive Monte Carlo simulations and real-world experiments in a semi-controlled environment (public swimming pool), employing two types of 3D complex trajectories (ellipse and Lissajous) in three-degrees-of-freedom tracking and two-degrees-of-stability scenarios. The proposed analytic control allocation solution (CA$_{prop}$) was found to be on par with state of the art allocation with enhanced computational efficiency in both the simulation and experimental settings, bringing us one step closer to realising fully autonomous 6-DOF navigation for fin-actuated AUVs.

Future research can improve real world performance of the proposed control framework. Specifically, we want to investigate input filters for the hybrid adaptive controller to avoid self-induced error oscillations such as shown in \cite{meger20143d}. Additionally, we would like to investigate the effectiveness of adaptively changing the number of involved fins for each DOF. We would also like to further investigate optimization based control allocation with better definitions of the cost function and the constraints.
Given the current control framework we want to explore robust and fault-tolerant control for fin-actuated AUVs. Moreover, we intend to conduct real-world experiments in diverse underwater environments to further validate and refine our approach under varying conditions and challenges. To this end, the control approach can be easily extended to estimate additive disturbances such as ocean currents as shown in \cite{basso2022global}. However, this would need to be combined with a state estimation filter for the internal fin-induced disturbances to avoid negative effects on the disturbance estimation and rejection. Another interesting avenue for adaptation are the parameters used in the fin model to map required thrusts to amplitudes, specifically in real world complex flows adapting those parameters could prove beneficial. In terms of extending the autonomy capabilities of fin driven vehicles a more sophisticated trajectory generation \cite{xanthidis2020navigation} can be implemented. To further optimise the efficiency and performance of control allocation for fin-actuated vehicles, the integration of advanced reinforcement learning techniques will also be explored. 


\bibliographystyle{IEEEtran}
\bibliography{IEEEabrv,mybibfile}

\begin{thebibliography}{10}
\providecommand{\url}[1]{#1}
\csname url@samestyle\endcsname
\providecommand{\newblock}{\relax}
\providecommand{\bibinfo}[2]{#2}
\providecommand{\BIBentrySTDinterwordspacing}{\spaceskip=0pt\relax}
\providecommand{\BIBentryALTinterwordstretchfactor}{4}
\providecommand{\BIBentryALTinterwordspacing}{\spaceskip=\fontdimen2\font plus
\BIBentryALTinterwordstretchfactor\fontdimen3\font minus \fontdimen4\font\relax}
\providecommand{\BIBforeignlanguage}[2]{{%
\expandafter\ifx\csname l@#1\endcsname\relax
\typeout{** WARNING: IEEEtran.bst: No hyphenation pattern has been}%
\typeout{** loaded for the language `#1'. Using the pattern for}%
\typeout{** the default language instead.}%
\else
\language=\csname l@#1\endcsname
\fi
#2}}
\providecommand{\BIBdecl}{\relax}
\BIBdecl

\bibitem{dudek2007aqua}
G.~Dudek, P.~Giguere, C.~Prahacs, S.~Saunderson, J.~Sattar, L.-A. Torres-Mendez, M.~Jenkin, A.~German, A.~Hogue, A.~Ripsman \emph{et~al.}, ``Aqua: An amphibious autonomous robot,'' \emph{Computer}, vol.~40, no.~1, pp. 46--53, 2007.

\bibitem{siegenthaler2013system}
C.~Siegenthaler, C.~Pradalier, F.~G{\"u}nther, G.~Hitz, and R.~Siegwart, ``System integration and fin trajectory design for a robotic sea-turtle,'' in \emph{2013 IEEE/RSJ International Conference on Intelligent Robots and Systems}.\hskip 1em plus 0.5em minus 0.4em\relax IEEE, 2013, pp. 3790--3795.

\bibitem{Salumae14}
T.~Salumae, R.~Raag, J.~Rebane, A.~Ernits, G.~Toming, M.~Ratas, and M.~Kruusmaa, ``The arrows project: adapting and developing robotics technologies for underwater archaeology,'' in \emph{IEEE Oceans-St. John's}, 2014, pp. 1--5.

\bibitem{basso2022global}
E.~A. Basso, H.~M. Schmidt-Didlaukies, K.~Y. Pettersen, and A.~J. Sorensen, ``Global asymptotic tracking for marine vehicles using adaptive hybrid feedback,'' \emph{IEEE Transactions on Automatic Control}, 2022.

\bibitem{konno2005development}
A.~Konno, T.~Furuya, A.~Mizuno, K.~Hishinuma, K.~Hirata, and M.~Kawada, ``Development of turtle-like submergence vehicle,'' in \emph{Proceedings of the 7th international symposium on marine engineering}, 2005.

\bibitem{low2007modular}
K.-H. Low, C.~Zhou, T.~Ong, and J.~Yu, ``Modular design and initial gait study of an amphibian robotic turtle,'' in \emph{2007 IEEE International Conference on Robotics and Biomimetics (ROBIO)}.\hskip 1em plus 0.5em minus 0.4em\relax IEEE, 2007, pp. 535--540.

\bibitem{zhao2008development}
W.~Zhao, Y.~Hu, L.~Wang, and Y.~Jia, ``Development of a flipper propelled turtle-like underwater robot and its cpg-based control algorithm,'' in \emph{2008 47th IEEE Conference on Decision and Control}.\hskip 1em plus 0.5em minus 0.4em\relax IEEE, 2008, pp. 5226--5231.

\bibitem{yao2013development}
G.~Yao, J.~Liang, T.~Wang, X.~Yang, Q.~Shen, Y.~Zhang, H.~Wu, and W.~Tian, ``Development of a turtle-like underwater vehicle using central pattern generator,'' in \emph{2013 IEEE International Conference on Robotics and Biomimetics (ROBIO)}.\hskip 1em plus 0.5em minus 0.4em\relax IEEE, 2013, pp. 44--49.

\bibitem{wang2012cpg}
C.~Wang, G.~Xie, X.~Yin, L.~Li, and L.~Wang, ``Cpg-based locomotion control of a quadruped amphibious robot,'' in \emph{2012 IEEE/ASME International Conference on Advanced Intelligent Mechatronics (AIM)}.\hskip 1em plus 0.5em minus 0.4em\relax IEEE, 2012, pp. 1--6.

\bibitem{geder2013maneuvering}
J.~D. Geder, R.~Ramamurti, M.~Pruessner, and J.~Palmisano, ``Maneuvering performance of a four-fin bio-inspired uuv,'' in \emph{2013 OCEANS-San Diego}.\hskip 1em plus 0.5em minus 0.4em\relax IEEE, 2013, pp. 1--7.

\bibitem{licht2008biomimetic}
S.~C. Licht, ``Biomimetic oscillating foil propulsion to enhance underwater vehicle agility and maneuverability,'' WOODS HOLE OCEANOGRAPHIC INSTITUTION MA, Tech. Rep., 2008.

\bibitem{chemori2016depth}
A.~Chemori, K.~Kuusmik, T.~Salum{\"a}e, and M.~Kruusmaa, ``Depth control of the biomimetic u-cat turtle-like auv with experiments in real operating conditions,'' in \emph{2016 IEEE International Conference on Robotics and Automation (ICRA)}.\hskip 1em plus 0.5em minus 0.4em\relax IEEE, 2016, pp. 4750--4755.

\bibitem{fischer2014nonlinear}
N.~Fischer, D.~Hughes, P.~Walters, E.~M. Schwartz, and W.~E. Dixon, ``Nonlinear rise-based control of an autonomous underwater vehicle,'' \emph{IEEE Transactions on Robotics}, vol.~30, no.~4, pp. 845--852, 2014.

\bibitem{plamondon2009trajectory}
N.~Plamondon and M.~Nahon, ``A trajectory tracking controller for an underwater hexapod vehicle,'' \emph{Bioinspiration \& biomimetics}, vol.~4, no.~3, p. 036005, 2009.

\bibitem{plamondon2010modeling}
N.~Plamondon, ``Modeling and control of a biomimetic underwater vehicle,'' Ph.D. dissertation, McGill University, 2010.

\bibitem{giguere2013wide}
P.~Giguere, Y.~Girdhar, and G.~Dudek, ``Wide-speed autopilot system for a swimming hexapod robot,'' in \emph{2013 International Conference on Computer and Robot Vision}.\hskip 1em plus 0.5em minus 0.4em\relax IEEE, 2013, pp. 9--15.

\bibitem{salumae2017motion}
T.~Salum{\"a}e, A.~Chemori, and M.~Kruusmaa, ``Motion control of a hovering biomimetic four-fin underwater robot,'' \emph{IEEE Journal of Oceanic Engineering}, vol.~44, no.~1, pp. 54--71, 2017.

\bibitem{fossen2011handbook}
T.~I. Fossen, \emph{Handbook of marine craft hydrodynamics and motion control}.\hskip 1em plus 0.5em minus 0.4em\relax John Wiley \& Sons, 2011.

\bibitem{meger20143d}
D.~Meger, F.~Shkurti, D.~C. Poza, P.~Giguere, and G.~Dudek, ``3d trajectory synthesis and control for a legged swimming robot,'' in \emph{2014 IEEE/RSJ International Conference on Intelligent Robots and Systems}.\hskip 1em plus 0.5em minus 0.4em\relax IEEE, 2014, pp. 2257--2264.

\bibitem{smallwood2002effect}
D.~A. Smallwood and L.~L. Whitcomb, ``The effect of model accuracy and thruster saturation on tracking performance of model based controllers for underwater robotic vehicles: experimental results,'' in \emph{Proceedings 2002 IEEE International Conference on Robotics and Automation (Cat. No. 02CH37292)}, vol.~2.\hskip 1em plus 0.5em minus 0.4em\relax IEEE, 2002, pp. 1081--1087.

\bibitem{fossen1991adaptive}
T.~I. Fossen and S.~I. Sagatun, ``Adaptive control of nonlinear systems: A case study of underwater robotic systems,'' \emph{Journal of Robotic Systems}, vol.~8, no.~3, pp. 393--412, 1991.

\bibitem{antonelli2001novel}
G.~Antonelli, F.~Caccavle, S.~Chiaverini, and G.~Fusco, ``A novel adaptive control law for autonomous underwater vehicles,'' in \emph{Proceedings 2001 ICRA. IEEE International Conference on Robotics and Automation (Cat. No. 01CH37164)}, vol.~1.\hskip 1em plus 0.5em minus 0.4em\relax IEEE, 2001, pp. 447--452.

\bibitem{von2018stable}
K.~D. von Ellenrieder, ``Stable backstepping control of marine vehicles with actuator rate limits and saturation,'' \emph{IFAC-PapersOnLine}, vol.~51, no.~29, pp. 262--267, 2018.

\bibitem{fresk2013full}
E.~Fresk and G.~Nikolakopoulos, ``Full quaternion based attitude control for a quadrotor,'' in \emph{2013 European control conference (ECC)}.\hskip 1em plus 0.5em minus 0.4em\relax IEEE, 2013, pp. 3864--3869.

\bibitem{louis2017quaternion}
S.~Louis, L.~Lapierre, Y.~Onmek, K.~G. Dejean, T.~Claverie, and S.~Vill{\'e}ger, ``Quaternion based control for robotic observation of marine diversity,'' in \emph{OCEANS 2017-Aberdeen}.\hskip 1em plus 0.5em minus 0.4em\relax IEEE, 2017, pp. 1--7.

\bibitem{bhat2000topological}
S.~P. Bhat and D.~S. Bernstein, ``A topological obstruction to continuous global stabilization of rotational motion and the unwinding phenomenon,'' \emph{Systems \& control letters}, vol.~39, no.~1, pp. 63--70, 2000.

\bibitem{fjellstad1994singularity}
O.-E. Fjellstad and T.~I. Fossen, ``Singularity-free tracking of unmanned underwater vehicles in 6 dof,'' in \emph{Proceedings of 1994 33rd IEEE Conference on Decision and Control}, vol.~2.\hskip 1em plus 0.5em minus 0.4em\relax IEEE, 1994, pp. 1128--1133.

\bibitem{mayhew2011quaternion}
C.~G. Mayhew, R.~G. Sanfelice, and A.~R. Teel, ``Quaternion-based hybrid control for robust global attitude tracking,'' \emph{IEEE Transactions on Automatic control}, vol.~56, no.~11, pp. 2555--2566, 2011.

\bibitem{teel2007robust}
A.~R. Teel, ``Robust hybrid control systems: An overview of some recent results,'' \emph{Advances in control theory and applications}, pp. 279--302, 2007.

\bibitem{johansen2013control}
T.~A. Johansen and T.~I. Fossen, ``Control allocation—a survey,'' \emph{Automatica}, vol.~49, no.~5, pp. 1087--1103, 2013.

\bibitem{durham1993constrained}
W.~C. Durham, ``Constrained control allocation,'' \emph{Journal of Guidance, control, and Dynamics}, vol.~16, no.~4, pp. 717--725, 1993.

\bibitem{oppenheimer2006control}
M.~W. Oppenheimer, D.~B. Doman, and M.~A. Bolender, ``Control allocation for over-actuated systems,'' in \emph{2006 14th Mediterranean Conference on Control and Automation}.\hskip 1em plus 0.5em minus 0.4em\relax IEEE, 2006, pp. 1--6.

\bibitem{adams2012robust}
R.~J. Adams, J.~M. Buffington, A.~G. Sparks, and S.~S. Banda, \emph{Robust multivariable flight control}.\hskip 1em plus 0.5em minus 0.4em\relax Springer Science \& Business Media, 2012.

\bibitem{buffington1996lyapunov}
J.~M. Buffington and D.~F. Enns, ``Lyapunov stability analysis of daisy chain control allocation,'' \emph{Journal of Guidance, Control, and Dynamics}, vol.~19, no.~6, pp. 1226--1230, 1996.

\bibitem{bodson2002evaluation}
M.~Bodson, ``Evaluation of optimization methods for control allocation,'' \emph{Journal of Guidance, Control, and Dynamics}, vol.~25, no.~4, pp. 703--711, 2002.

\bibitem{paradiso1991adaptable}
J.~A. Paradiso, ``Adaptable method of managing jets and aerosurfaces for aerospace vehicle control,'' \emph{Journal of Guidance, Control, and Dynamics}, vol.~14, no.~1, pp. 44--50, 1991.

\bibitem{harkegard2002efficient}
O.~Harkegard, ``Efficient active set algorithms for solving constrained least squares problems in aircraft control allocation,'' in \emph{Proceedings of the 41st IEEE Conference on Decision and Control, 2002.}, vol.~2.\hskip 1em plus 0.5em minus 0.4em\relax IEEE, 2002, pp. 1295--1300.

\bibitem{petersen2005constrained}
J.~A. Petersen and M.~Bodson, ``Constrained quadratic programming techniques for control allocation,'' \emph{IEEE Transactions on Control Systems Technology}, vol.~14, no.~1, pp. 91--98, 2005.

\bibitem{jin2015six}
S.~Jin, J.~Kim, J.~Kim, and T.~Seo, ``Six-degree-of-freedom hovering control of an underwater robotic platform with four tilting thrusters via selective switching control,'' \emph{IEEE/ASME Transactions on mechatronics}, vol.~20, no.~5, pp. 2370--2378, 2015.

\bibitem{bak2022hovering}
J.~Bak, Y.~Moon, J.~Kim, S.~Mohan, T.~Seo, and S.~Jin, ``Hovering control of an underwater robot with tilting thrusters using the decomposition and compensation method based on a redundant actuation model,'' \emph{Robotics and Autonomous Systems}, vol. 150, p. 103995, 2022.

\bibitem{breivik2009guidance}
M.~Breivik and T.~I. Fossen, ``Guidance laws for autonomous underwater vehicles,'' \emph{Underwater vehicles}, vol.~4, pp. 51--76, 2009.

\bibitem{remmas2021inverse}
W.~Remmas, A.~Chemori, and M.~Kruusmaa, ``Inverse-model intelligent control of fin-actuated underwater robots based on drag force propulsion,'' \emph{Ocean Engineering}, vol. 239, p. 109883, 2021.

\bibitem{sanfelice2006robust}
R.~G. Sanfelice, M.~J. Messina, S.~E. Tuna, and A.~R. Teel, ``Robust hybrid controllers for continuous-time systems with applications to obstacle avoidance and regulation to disconnected set of points,'' in \emph{2006 American Control Conference}.\hskip 1em plus 0.5em minus 0.4em\relax IEEE, 2006, pp. 6--pp.

\bibitem{goebel2009hybrid}
R.~Goebel, R.~G. Sanfelice, and A.~R. Teel, ``Hybrid dynamical systems,'' \emph{IEEE control systems magazine}, vol.~29, no.~2, pp. 28--93, 2009.

\bibitem{slotine1990hamiltonian}
J.~Slotine and M.~Di~Benedetto, ``Hamiltonian adaptive control of spacecraft,'' \emph{IEEE Transactions on Automatic Control}, vol.~35, no.~7, pp. 848--852, 1990.

\bibitem{paden1988globally}
B.~Paden and R.~Panja, ``Globally asymptotically stable ‘pd+’controller for robot manipulators,'' \emph{International Journal of Control}, vol.~47, no.~6, pp. 1697--1712, 1988.

\bibitem{krstic1995nonlinear}
M.~Krstic, P.~V. Kokotovic, and I.~Kanellakopoulos, \emph{Nonlinear and adaptive control design}.\hskip 1em plus 0.5em minus 0.4em\relax John Wiley \& Sons, Inc., 1995.

\bibitem{ren2015hydrodynamic}
Z.~Ren, T.~Wang, and L.~Wen, ``Hydrodynamic function of a robotic fish caudal fin: effect of kinematics and flow speed,'' in \emph{2015 IEEE/RSJ International Conference on Intelligent Robots and Systems (IROS)}.\hskip 1em plus 0.5em minus 0.4em\relax IEEE, 2015, pp. 3882--3887.

\bibitem{fossen2006survey}
T.~I. Fossen and T.~A. Johansen, ``A survey of control allocation methods for ships and underwater vehicles,'' in \emph{2006 14th Mediterranean Conference on Control and Automation}.\hskip 1em plus 0.5em minus 0.4em\relax IEEE, 2006, pp. 1--6.

\bibitem{sqp}
P.~T. Boggs and J.~W. Tolle, ``Sequential quadratic programming,'' \emph{Acta numerica}, vol.~4, pp. 1--51, 1995.

\bibitem{bracewellHeaviside}
R.~Bracewell, \emph{The Fourier Transform and its Applications, 3rd ed.}\hskip 1em plus 0.5em minus 0.4em\relax New York: McGraw-Hill, 2000.

\bibitem{sproewitz2008learning}
A.~Sproewitz, R.~Moeckel, J.~Maye, and A.~J. Ijspeert, ``Learning to move in modular robots using central pattern generators and online optimization,'' \emph{The International Journal of Robotics Research}, vol.~27, no. 3-4, pp. 423--443, 2008.

\bibitem{georgiades2009simulation}
C.~Georgiades, M.~Nahon, and M.~Buehler, ``Simulation of an underwater hexapod robot,'' \emph{Ocean Engineering}, vol.~36, no.~1, pp. 39--47, 2009.

\bibitem{healey1995toward}
A.~J. Healey, S.~Rock, S.~Cody, D.~Miles, and J.~Brown, ``Toward an improved understanding of thruster dynamics for underwater vehicles,'' \emph{IEEE Journal of oceanic Engineering}, vol.~20, no.~4, pp. 354--361, 1995.

\bibitem{georgiades2004aqua}
C.~Georgiades, A.~German, A.~Hogue, H.~Liu, C.~Prahacs, A.~Ripsman, R.~Sim, L.-A. Torres, P.~Zhang, M.~Buehler \emph{et~al.}, ``Aqua: an aquatic walking robot,'' in \emph{2004 IEEE/RSJ International Conference on Intelligent Robots and Systems (IROS)(IEEE Cat. No. 04CH37566)}, vol.~4.\hskip 1em plus 0.5em minus 0.4em\relax IEEE, 2004, pp. 3525--3531.

\bibitem{manhaes2017metrics}
M.~M.~M. Manhaes, S.~A. Scherer, L.~R. Douat, M.~Voss, and T.~Rauschenbach, ``Use of simulation-based performance metrics on the evaluation of dynamic positioning controllers,'' in \emph{OCEANS 2017-Aberdeen}.\hskip 1em plus 0.5em minus 0.4em\relax IEEE, 2017, pp. 1--8.

\bibitem{genetic2020}
R.~M. Solgi, ``geneticalgorithm,'' \url{https://pypi.org/project/geneticalgorithm/}, 2020, accessed: March 27, 2023.

\bibitem{garrido2014automatic}
S.~Garrido-Jurado, R.~Mu{\~n}oz-Salinas, F.~J. Madrid-Cuevas, and M.~J. Mar{\'\i}n-Jim{\'e}nez, ``Automatic generation and detection of highly reliable fiducial markers under occlusion,'' \emph{Pattern Recognition}, vol.~47, no.~6, pp. 2280--2292, 2014.

\bibitem{ALGLIB}
A.~Project, ``Alglib,'' \url{https://www.alglib.net/}, 2023, accessed: March 27, 2023.

\bibitem{salumae2016motion}
T.~Salum{\"a}e, A.~Chemori, and M.~Kruusmaa, ``Motion control architecture of a 4-fin u-cat auv using dof prioritization,'' in \emph{2016 IEEE/RSJ International Conference on Intelligent Robots and Systems (IROS)}.\hskip 1em plus 0.5em minus 0.4em\relax IEEE, 2016, pp. 1321--1327.

\bibitem{xanthidis2020navigation}
M.~Xanthidis, N.~Karapetyan, H.~Damron, S.~Rahman, J.~Johnson, A.~O’Connell, J.~M. O’Kane, and I.~Rekleitis, ``Navigation in the presence of obstacles for an agile autonomous underwater vehicle,'' in \emph{2020 IEEE International Conference on Robotics and Automation (ICRA)}.\hskip 1em plus 0.5em minus 0.4em\relax IEEE, 2020, pp. 892--899.

\bibitem{fjellstad1994quaternion}
Fjellstad and Fossen, ``Quaternion feedback regulation of underwater vehicles,'' in \emph{1994 Proceedings of IEEE International Conference on Control and Applications}.\hskip 1em plus 0.5em minus 0.4em\relax IEEE, 1994, pp. 857--862.

\bibitem{palomeras2013}
N.~Palomeras, S.~Nagappa, D.~Ribas, N.~Gracias, and M.~Carreras, ``Vision-based localization and mapping system for auv intervention,'' in \emph{2013 MTS/IEEE OCEANS-Bergen}.\hskip 1em plus 0.5em minus 0.4em\relax IEEE, 2013, pp. 1--7.

\end{thebibliography}
\begin{IEEEbiography}[{\includegraphics[width=1in,height=1.2in,clip,keepaspectratio]{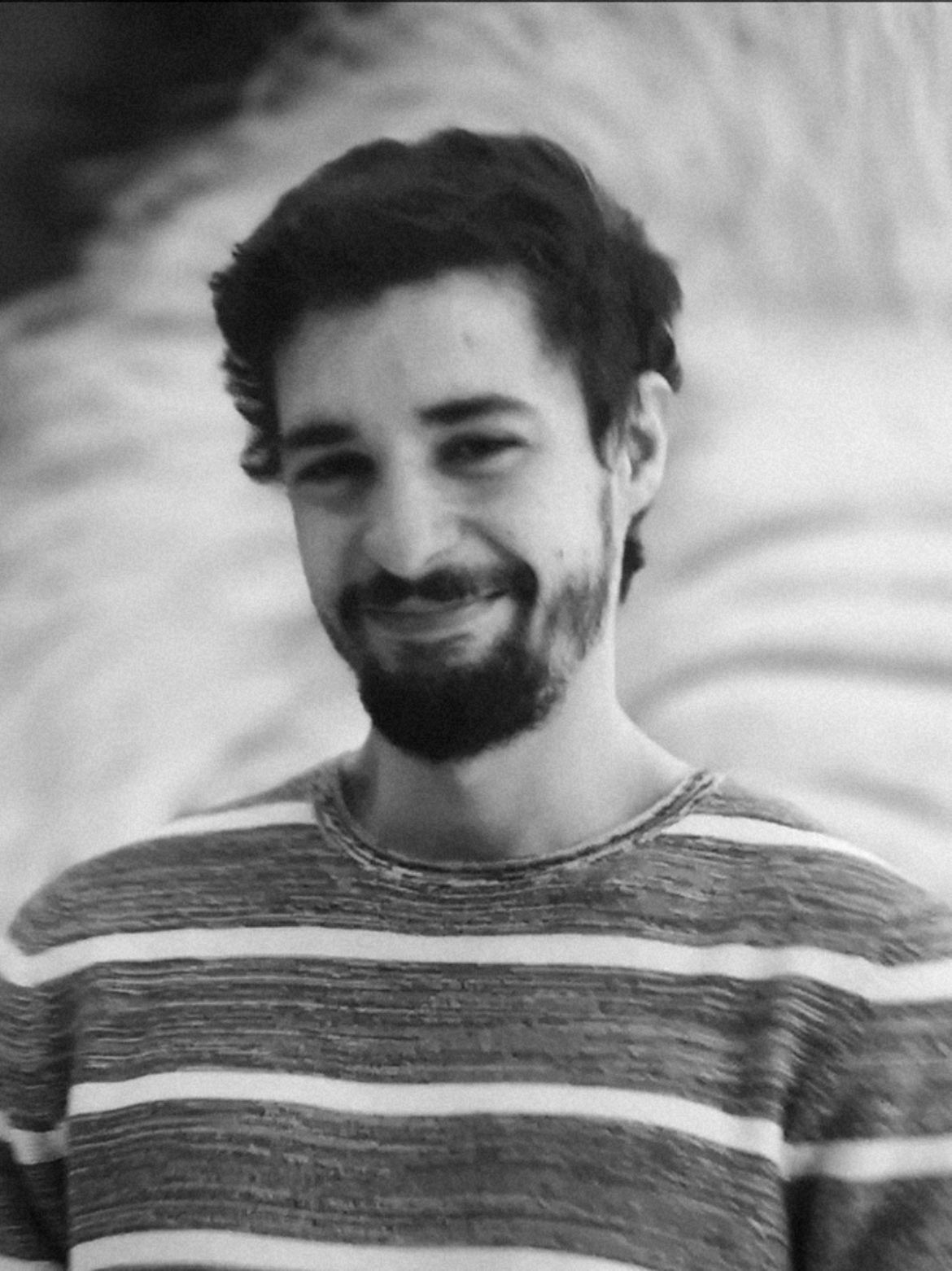}}]{Walid Remmas}
received his M.Sc. in Control System Engineering from the Polytechnic School of Constantine, Algeria, in 2017. He then went on to earn a second M.Sc. in Robotics from the University of Montpellier, France, in 2018. In 2023, he was awarded a joint Ph.D. in Computer and Systems Engineering from Tallinn University of Technology, Estonia, and the University of Montpellier, France. He is currently a researcher at Defsecintel Solutions, where he continues to contribute to advancements in computer vision and control. His research interests include intelligent control, underwater robotics, and computer vision.
\end{IEEEbiography}

\begin{IEEEbiography}[{\includegraphics[width=1in,height=1.2in,clip,keepaspectratio]{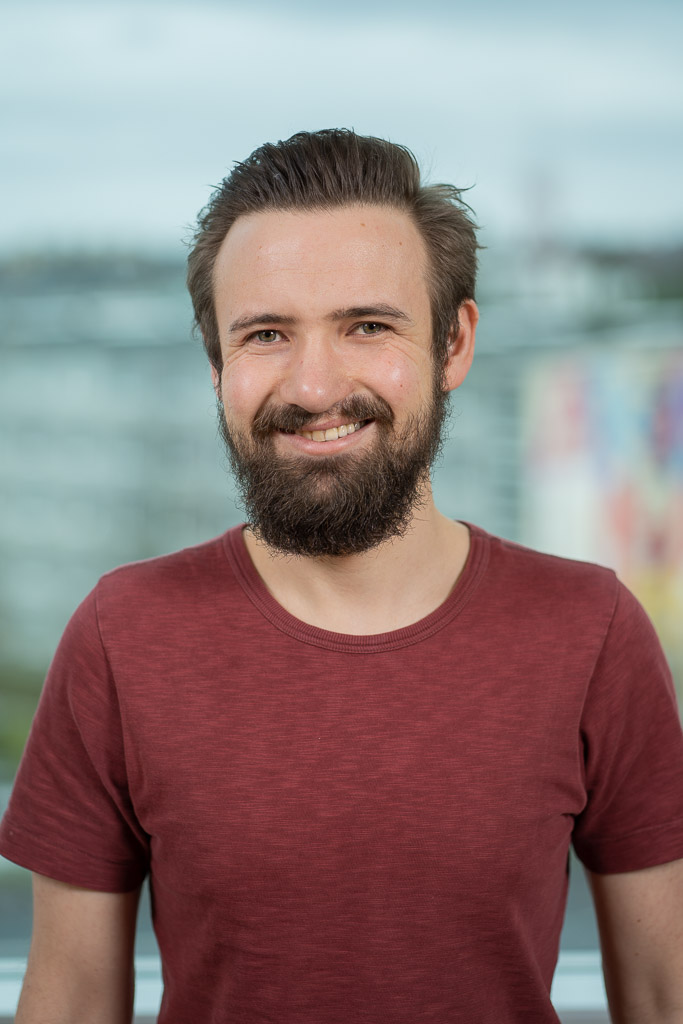}}]{Christian Meurer}
received his Ph.D. in Computer and Systems Engineering from Tallinn University of Technology (TalTech), Tallinn, Estonia in 2021. He is currently working as postdoctoral researcher at the Center for Marine Environmental Sciences (MARUM), University of Bremen, Bremen, Germany. His research interests include bio-inspired sensing and actuation for underwater vehicles, as well as, nonlinear modeling and control.  
\end{IEEEbiography}

\begin{IEEEbiography}[{\includegraphics[width=1in,height=1.2in,clip,keepaspectratio]{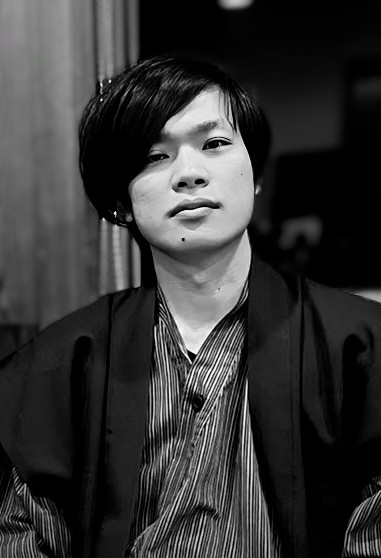}}]{Yuya Hamamatsu} received his M.Sc. degree in environment from the University of Tokyo, Japan, in 2020. He is currently working toward the Ph.D. degree with the Centre for Biorobotics at Tallinn University of Technology, Estonia. His research interests control theory on robotics.
\end{IEEEbiography}

\begin{IEEEbiography}[{\includegraphics[width=1in,height=1.2in,clip,keepaspectratio]{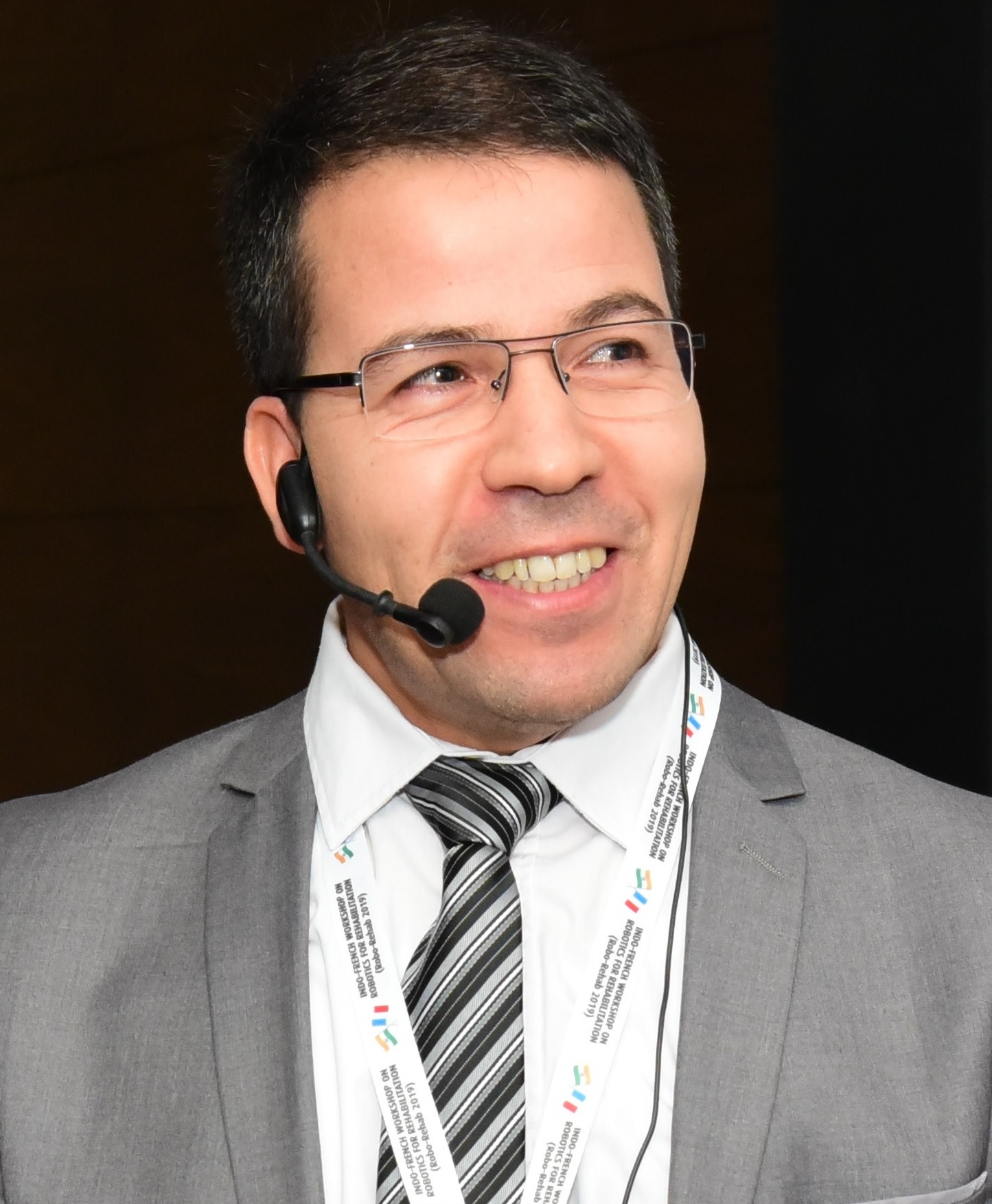}}]{Ahmed Chemori}
 received the M.Sc. and Ph.D. degrees both in automatic control from the Grenoble Institute of Technology, Grenoble, France, in 2001 and 2005, respectively. He has been a Postdoctoral Fellow with the Automatic Control Laboratory, Grenoble, France, in 2006. He is currently a tenured Research Scientist in automatic control and robotics with the Montpellier Laboratory of Informatics, Robotics, and Microelectronics. His research interests include nonlinear robust adaptive, and predictive control and their real-time applications in robotics.
\end{IEEEbiography}

\begin{IEEEbiography}[{\includegraphics[width=1in,height=1.2in,clip,keepaspectratio]{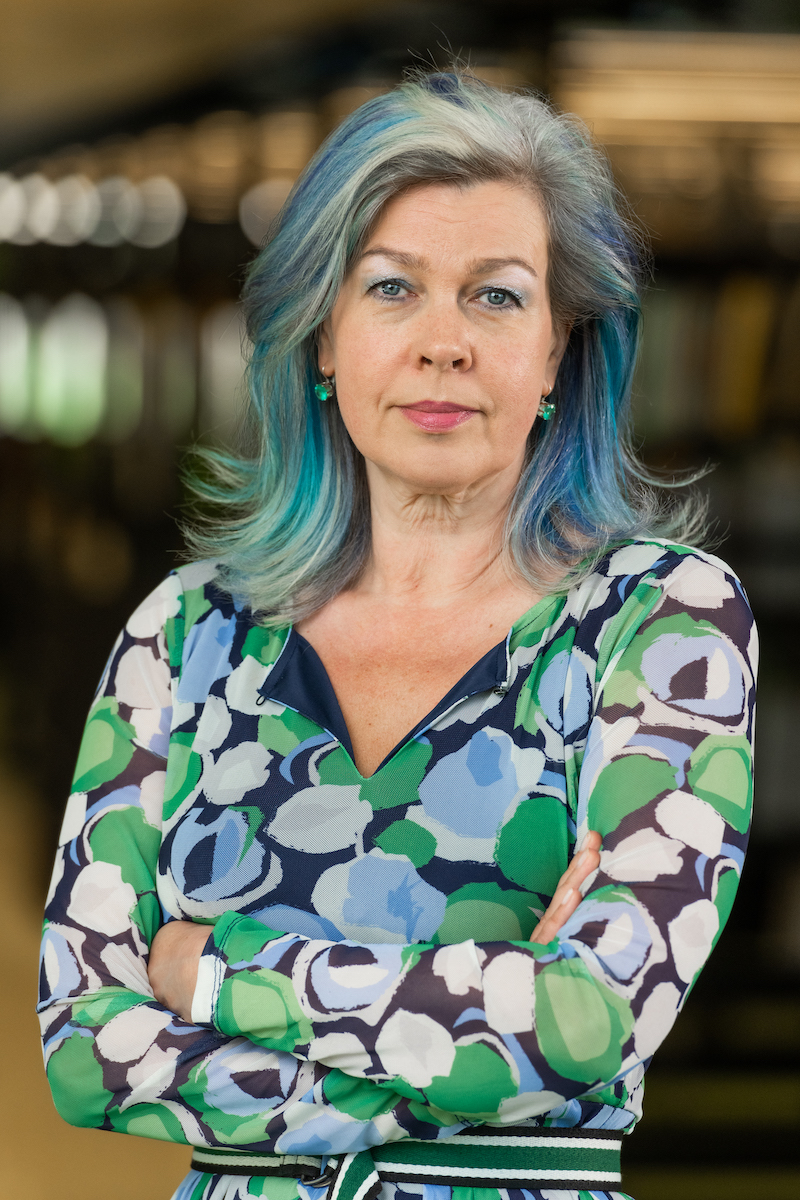}}]{Maarja Kruusmaa}
 received her Ph.D.from Chalmers University of Technology in 2002 and since 2008 is a professor in Tallinn University of Technology (TalTech) as a PI of Centre for Biororobitcs, a research group focusing on bio-inspired robotics, underwater robotics and novel underwater sensing technologies. 2017 - 2022 she was a visiting professor in NTNU AMOS (Centre for Excellence of Autonomous Marine Operations and Systems). Her research interests include novel locomotion mechanisms for underwater environments and on flowable media and novel methods for underwater flow sensing. 
\end{IEEEbiography}


\appendices
\section{ Kinematic and Dynamic Model}
\label{sec:appendix1}

Quaternions are an efficient orientation representation for movements in 6-DOF \cite{fjellstad1994quaternion}. More specifically, unit quaternions $q \in \mathbb{S}^3$, having the property $\lVert q \rVert = 1$ are used. They can be represented as a vector $q = (\mu, \epsilon)$ constituting the quaternion's scalar $\mu \in \mathbb{R}$ and vector $\epsilon \in \mathbb{R}^3$ parts. Note that we will follow the Hamiltonian notation for quaternions in this paper. 
We define the inverse of a unit quaternion as it's complex conjugate: 

\begin{equation}
    q^{-1} = \bar{q} = 
    \begin{bmatrix}
        \mu \\
        -\epsilon
    \end{bmatrix}
    \in \mathbb{S}^3
\end{equation}

and the unit quaternion's identity as:

\begin{equation}
    1_q = 
    \begin{bmatrix}
        1 \\
        0_{3 \times 1}
    \end{bmatrix}
    \in \mathbb{S}^3
\end{equation}

Furthermore, we can define quaternion multiplication as: 

\begin{equation}\label{eq:quatmult}
    q_1 \odot q_2 = 
    \begin{bmatrix}
        \mu_1 \mu_2 - \epsilon_1^T \epsilon_2 \\
        \mu_1 \epsilon_2 + [\epsilon_1]_{\times} \epsilon_2
    \end{bmatrix}
\end{equation}

where $[\cdot]_{\times} \in \mathbb{R}^{3 \times 3}$, denotes the skew symmetric matrix representation

The position of an underwater vehicle in the three dimensional space can be uniquely defined by a vector $p \in \mathbb{R}^3$ describing the origin of the body fixed frame with respect to a fixed inertial frame. The attitude of an underwater vehicle in three dimensional space can be represented by a rotation matrix $R \in SO(3)$ mapping the axes of the body fixed frame onto those of the inertial frame. The vehicles kinematics can then be written by, using the linear $v \in \mathbb{R}^3$ and angular $w \in \mathbb{R}^3$ velocities expressed in body fixed frame, in the following form:
\begin{equation}\label{eq:kinematicsSO3}
    \begin{split}
        \dot{p} &= Rv \\
        \dot{R} &= R[w]_{\times}
    \end{split}
\end{equation}

For efficient and intuitive representation the rotation matrix can be represented by various three-parameter parametrizations such as Tate-Brian or Euler angles \cite{fossen1991adaptive}. However, none of those parametrizations is globally non-singular, which makes them less conducive for use of 6-DOF vehicle control. An alternative solution is the use of the unit quaternion \cite{fjellstad1994singularity}.
The map from a unit quaternion to a rotation matrix $R(q): \mathbb{S}^3 \to SO(3)$ is described by: 
\begin{equation}\label{eq:S3ToSO3}
    R(q) := I_3 + 2\mu [\epsilon]_\times + 2[\epsilon]_{\times}^2
\end{equation}

Based on the definition in \eqref{eq:quatmult} $R$ is a group homorphism satisfying: 
\begin{equation}
    R(q_1) R(q_2) = R(q_1 \odot q_2)
\end{equation}
and additionally $R^{-1}(q) = R^T(q) = R(q^{-1})$, as well as $R(1_q) = R(-1_q) = I_{3}$

Subsequently, the kinematic equations can be rewritten as:
\begin{equation}
    \begin{split}
        \dot{p} &= R(q) v \\
        \dot{q} &= \frac{1}{2} q \odot \chi(w) = T(q) w
    \end{split}
\end{equation}

with $\chi: \mathbb{R}^3 \to \mathbb{R}^4$ defined as: 
\begin{equation}
    \chi(w) = \begin{bmatrix}
        0 \\
        w
    \end{bmatrix}
\end{equation}
and $T(q)$ as: 
\begin{equation}
    T(q) = \frac{1}{2} \begin{bmatrix}
        -\epsilon^T \\
        \mu I_3 + [\epsilon]_{\times}
    \end{bmatrix}
\end{equation}

The 6-DOF kino-dynamic model, expressed in body frame, can then be formulated following Fossen's vectorial notation \cite{fossen2011handbook} as:
\begin{equation}\label{eq:dynQuat} 
    \begin{split}
        M\dot{\nu}+C(\nu)\nu+D(\nu)\nu+g(q)  =  \tau \\
        \dot{\eta}  =  J(q)\nu
    \end{split}
\end{equation}

Where $\tau \in \mathbb{R}^6$ is the vector of wrenches acting as control input, while $\eta = [p, q]^T \in \mathbb{R}^7, \nu = [v, w]^T \in \mathbb{R}^6$ represent the vectors of the vehicle poses in the earth-fixed frame $R_n$ and the velocities in the body-fixed frame $R_b$ respectively. The Jacobian $J(q)$ combines the linear and angular kinematic mappings in the following way: 
\begin{equation}
    J(q) = 
    \begin{bmatrix}
        R(q) & 0_{3x3} \\
        0_{4 \times 3} & T(q)
    \end{bmatrix}
\end{equation}
with it's pseudoinverse: 
\begin{equation}\label{eq:TransformationQuatInv}
    J(q)^\dagger = \begin{bmatrix}
        R(q)^T & 0_{3x4} \\
        0_{3 \times 3} & 4T(q)^T
    \end{bmatrix}
\end{equation}


\section{Trajectory generation}
\label{sec:appendix2}
To test our proposed control framework we use two distinct trajectory tracking scenarios, full 6-DOF trajectory tracking (6T) and 3-DOF trajectory tracking (surge, heave, yaw) with roll and pitch stabilization (3T2S). In the 6T scenario we assumed that the robot can generate forces in all 6-DOF, while in the 3T2S scenario we assumed, that actuation in the sway direction is not efficient enough to be meaningful. The lack of available sway force makes the system non-holonomic and we implemented a look-ahead modification based on the respective state of the robot to generate a yaw trajectory, which ensured that the robot could still follow the desired trajectories without the need for sway forces. Within those two scenarios two types of looped analytic trajectories were tested. One trajectory is ellipsoidal, whereas the other one describes a Lissajous infinity figure. 

In the 6T scenario for both trajectory types the roll trajectory prescribes a motion at a constant angular velocity, while the pitch trajectory points the surge axis of the robot along the generated positional trajectory in the vertical plane. In contrast, in the 3T2S scenario roll and pitch are to be stabilized at zero. The analytic description for both trajectory types in the two scenarios can be seen in Table \ref{tab:AnalyticTrajectories}. In the given equations, $A_{x,y,z}$ describe the amplitudes of the looped trajectories in surge, sway and heave. The angular frequencies $\omega_{x, y, z}$ correlate with the desired velocities along the linear DOF, while $p_0 = [x_0, y_0, z_0]^T$ describes the starting position of the trajectory. For the roll DOF $c_{\phi}$ describes the roll coefficient, which specifies the desired roll velocity. Furthermore,  the look-ahead time $t^*$ and $\delta p_p(t, t^*) = p_p(t +t^*) - p_p(t)$ with $p_p = [y_p, y_p, z_p]^T$ are used to define yaw and pitch trajectories that point the robot's orientation along the positional trajectories in the horizontal and vertical plane respectively. 
\begin{table}[]

	\centering
	\vspace{1.5mm}
	\setlength\doublerulesep{0.5pt}
	\caption{Ellipsoid and Lissajous trajectory functions for 6T and 3T2S scenarios.}
	{
    \tiny
	\begin{tabularx}{0.5\textwidth}{>{\centering\arraybackslash}X|>{\centering\arraybackslash}X|>{\centering\arraybackslash}X|>{\centering\arraybackslash}X}
		& \textbf{DOF} & \textbf{Scenario 6T} & \textbf{Scenario 3T2S}  \\ \hhline{=|=|=|=}
		\multirow{6}{*}{\textbf{Ellipsoid}} & $x_p(t)$ & \multicolumn{2}{c}{$A_x (-\cos{(\omega_x t)} + 1) + x_0 $} \\
		                    & $y_p(t)$       & \multicolumn{2}{c}{$ A_y (\sin{(\omega_y t)}) + y_0$} \\ 
		                  & $z_p(t)$         & \multicolumn{2}{c}{$ A_z (-\cos{(\omega_z t)} + 1) + z_0 $} \\
		                  & $\phi_p(t)$         & $c_\phi t$ & $0$ \\
		                  & $\theta_p(t)$         & $\frac{\pi}{2} - \arccos{(\frac{-\delta z_p(t, t^*)}{||\delta p_p(t, t^*)||})}$ & $0$ \\
		                  & $\psi_p(t)$         & $ atan2(\delta y_p(t, t^*), \delta x_p(t, t^*))$ & $atan2(y_p(t+t^*) - y, ~ x_p(t+t^*) - x)$ \\ \cline{1-4}
		\multirow{6}{*}{\textbf{Lissajous}} & $x_p(t)$ & \multicolumn{2}{c}{$A_x (-\cos{(l_x \omega_x t)} + 1) + x_0$} \\
		                  &$y_p(t)$         & \multicolumn{2}{c}{$A_y (\sin{(l_y \omega_y t)}) + y_0$} \\
		                  &$z_p(t)$         & \multicolumn{2}{c}{$A_z(-\cos{(\omega_z t)} + 1) + z_0$} \\
		                  &$\phi_p(t)$         & $c_\phi t$ & $0$ \\
		                  &$\theta_p(t)$         & $\frac{\pi}{2} - \arccos{(\frac{-\delta z_p(t, \zeta)}{||\delta p_p(t, \zeta)||})}$ & $0$ \\
		                  &$\psi_p(t)$         & $atan2(\delta y_p(t, t^*), \delta x_p(t, t^*))$ & $atan2(\delta y_p(t+t^*), \delta x_p(t+t^*))$
	\end{tabularx}}
	\label{tab:AnalyticTrajectories}
\end{table}

The analytic trajectories $\eta_p = [x_p, y_p, z_p, \phi_p, \theta_p, \psi_p]^T$ are then filtered through a pair of second order ordinary differential equations (ODEs) producing desired positions $\eta_d = [x_d, y_d, z_d, \phi_d, \theta_p \psi_d]^T$, velocities $\dot{\eta}_d = [\dot{x}_d, \dot{y}_d, \dot{z}_d, \dot{\phi}_d, \dot{\theta}_d, \dot{\psi}_d]^T$ and accelerations $\ddot{\eta}_d = [\ddot{x}_d, \ddot{y}_d, \ddot{z}_d, \ddot{\phi}_d, \ddot{\theta}_d, \ddot{\psi}_d]^T$.


\begin{align}
    \ddot{\eta}_{d_1} + 2\gamma_1 \dot{\eta}_{d_1} &= \gamma_1^2 (\eta_{d_1} - \eta_p) \\
    \ddot{\eta}_{d} + 2\gamma_2 \dot{\eta}_{d} &= \gamma_2^2 (\eta_{d} - \eta_{d_1})
\end{align}

The ODE filters are implemented to guarantee the generation of smooth, continuous, and feasible velocities and accelerations, even in the presence of non-linearity, potentially induced by the lookahead components in yaw and pitch,  in the desired set-points. The parameters $\gamma_1$ and $\gamma_2$ have been selected manually to guarantee feasible accelerations by the robot. A double Euler integration is then performed to get the desired states $\eta_d$, $\dot{\eta}_d$ and $\ddot{\eta}_d$. To conform with the 6-DOF tracking control framework the desired orientation represented by the Euler angles $[\phi_d, \theta_d, \psi_d]^T$ as well as their derivatives $[\dot{\phi}_d, \dot{\theta}_d, \dot{\psi}_d]^T$ and $[\ddot{\phi}_d, \ddot{\theta}_d, \ddot{\psi}_d]^T$ are transformed into unit quaternions \cite{fossen2011handbook}.

\section{State estimation}
\label{sec:appendix3}
To provide the control framework with reliable information about the robot's state, we are using an EKF as presented in \cite{palomeras2013}. Position in the inertial frame $p \in \mathbb{R}^3$ and linear velocities $v \in \mathbb{R}^3$  are the states estimated by the filter $\xi = [p, v]^T \in \mathbb{R}^6$. A kinematic vehicle model assuming constant velocity is used in the filters prediction. The constant velocity assumption is justified for slow moving vehicles, and thus applicable in our scenario with demanded surge velocities not exceeding \SI{0.2}{m/s}. Orientation information is assumed to be an input to the model. The filter prediction at time step $k$ can be then described by: 

\begin{equation}\label{eq:ekfpredmodel}
    \xi_k^- = 
    \begin{bmatrix}
        p_{k-1} + {R}(q_k) \left(v_{k-1}~t +  n_{k-1} ~ \frac{t^2}{2} \right) \\
        v + n^{acc}_{k-1} ~ t \\
    \end{bmatrix}
\end{equation}

with $n^{acc} \in \mathbb{R}^3$ representing zero-mean white Gaussian acceleration noise. The acceleration noise covariances are then represented by the system noise covariance matrix $Q = diag(\sigma^2_{n^{acc}}) \in \mathbb{R}^{3 \times 3}$

and the prediction covariance:
\begin{equation}
    P^-_k = A^{EKF}_k P_{k-1} {A^{EKF}_k}^T + W^{EKF}_k Q^{EKF} {W^{EKF}_{k-1}}^T
\end{equation}

with $A^{EKF}$ and $W^{EKF}$ being Jacobians of \eqref{eq:ekfpredmodel} with respect to the filter states and system noises respectively. 
The prediction is followed by the standard EKF correction step:
\begin{equation}
    \begin{split}
        K_k &= P^-_k H^T(HP_k^-H^T + R^{EKF})^{-1} \\
        \xi_k &= \xi_k^- + K_k(z_k - H_k \xi_k^-) \\
        P_k &= (I - K_k H_k)P_k^-
    \end{split}
\end{equation}

using asynchronous measurement updates $z_k = H_k \xi_k + s_k$, which make use of a variable size allocation for the observation matrix $H_k$ as shown in Palomeras et al. \cite{palomeras2013} and where $s_k$ is the measurement noise. In the specific application here we have $H_k = 1$ to accommodate depth estimates from a pressure sensor and $H_k = I^{2x2}$ to accommodate planar position estimates from a camera based aruco tag detection. Additionally, the standard notation applies with $K_k$ being the Kalman gain and $R^{EKF}$ describing the measurement covariance matrix.  
\end{document}